\pdfoutput=1

\documentclass[11pt]{article}

\usepackage{acl}

\usepackage{times}
\usepackage{latexsym}

\usepackage[T1]{fontenc}

\usepackage[utf8]{inputenc}

\usepackage{microtype}
\usepackage{inconsolata}
\usepackage{graphicx}
\usepackage{amsmath}

\usepackage{multirow}
\usepackage{booktabs}
\usepackage{subcaption}
\usepackage{amsmath}
\usepackage{amssymb}
\usepackage{amsfonts}
\usepackage{colortbl}
\usepackage{xcolor}
\usepackage[normalem]{ulem}
\usepackage{amsmath}
\usepackage{algorithmic}
\usepackage{algorithm}
\usepackage{fdsymbol}
\usepackage{paralist}

\title{Pruning Multilingual Large Language Models for Multilingual Inference}

\author{
    Hwichan Kim$^1$\quad Jun Suzuki$^2$ \quad Tosho Hirasawa$^1$ \quad Mamoru Komachi$^3$ \\
    $^1$Tokyo Metropolitan University, $^2$Tohoku University, $^3$Hitotsubashi University\\
    {\normalsize \texttt{\{kim-hwichan@ed.tmu, jun.suzuki@tohoku, toshosan@tmu, mamoru.komachi@hit-u\}.ac.jp}}
}

\begin{document}
\maketitle
\begin{abstract}
Multilingual large language models (MLLMs), trained on multilingual balanced data, demonstrate better zero-shot learning performance in non-English languages compared to large language models trained on English-dominant data. 
However, the disparity in performance between English and non-English languages remains a challenge yet to be fully addressed.
A distinctive characteristic of MLLMs is their high-quality translation capabilities, indicating an acquired proficiency in aligning between languages. 
This study explores how to enhance the zero-shot performance of MLLMs in non-English languages by leveraging their alignment capability between English and non-English languages.
To achieve this, we first analyze the behavior of MLLMs when performing translation and reveal that there are large magnitude features that play a critical role in the translation process. 
Inspired by these findings, we retain the weights associated with operations involving the large magnitude features and prune other weights to force MLLMs to rely on these features for tasks beyond translation.
We empirically demonstrate that this pruning strategy can enhance the MLLMs' performance in non-English language.\footnote{The code used in our experiments is available at \url{https://github.com/hwichan0720/pruning_for_multilinguality}.}

\end{abstract}

\section{Introduction}
\label{sec:intro}
Large Language Models (LLMs) have demonstrated remarkable language reasoning capabilities, particularly in English contexts. 
However, these LLMs limited their proficiency in non-English language \cite{ahuja-etal-2023-mega, ahuja2024megaverse}.
Recognizing this limitation, recent research endeavors have given rise to Multilingual Large Language Models (MLLMs), such as XGLM \cite{lin-etal-2022-shot}, mGPT \cite{10.1162/tacl_a_00633}, and BLOOM \cite{bigscience_workshop_2022}, designed to address the challenges posed by linguistic diversity. 
In the context of zero-shot learning, MLLMs have demonstrated superior proficiency across multiple languages compared to English LLMs \cite{etxaniz2023multilingual}.
Nevertheless, a discernible disparity persists in accuracy levels when comparing results between English and non-English languages.
For example, in the Cross-lingual NLI (XNLI) task \cite{conneau-etal-2018-xnli}, XGLM-2.9B achieves accuracies of 51.1 and 39.2 in English and Russian, respectively (see Tab. \ref{table:zero-shot}).

\begin{figure}[t]
 \centering
 \includegraphics[scale=0.37]{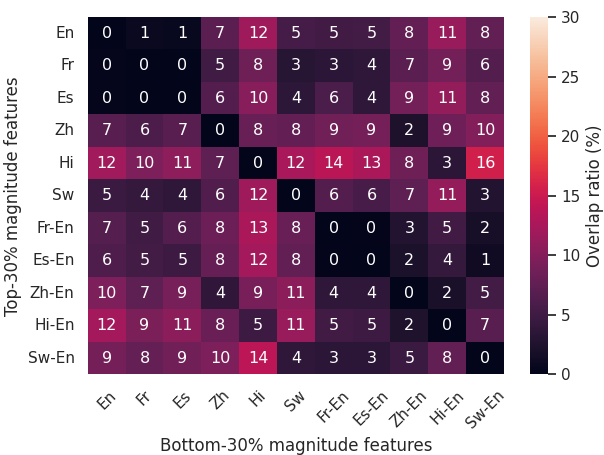}
 \caption{The overlap ratios among the top- and bottom-30\% features in the 27-th layer of XGLM, ranked by their magnitude. The row and column labels correspond to languages and language pairs used in few-shot monolingual (En, Fr, etc.) and translation (Fr-En, Es-En, etc.) demonstrations, respectively. Each element represents the ratios of overlapping features between the top- and bottom-30\% in magnitude within each demonstration. 
 This figure shows that specific features are active only when inputting translation demonstrations.}\label{fig:overlap_layer27}
\end{figure}

Achieving performance parity in non-English languages with English, which represents the upper bound of MLLMs, necessitates the alignment of English and non-English texts. 
\citet{ahuja-etal-2023-mega} has demonstrated the effectiveness of the translate-test, a methodology translating non-English texts into English using an external machine translation (MT) system and running inference over the translated texts, which serves as an approach to superficially align English and non-English.
However, the translate-test approach increases inference costs due to its reliance on the MT system. 
Conversely, recent research indicated that MLLMs can perform translation by incorporating few-shot translation demonstrations into the contextual information \cite{lin-etal-2022-shot,etxaniz2023multilingual,vilar-etal-2023-prompting}.
These findings suggest that MLLMs already have alignment capability between English and non-English, and it is manifested through the incorporation of few-shot translation demonstrations.
We believe that if the alignment capability can be brought out in tasks other than translation, non-English performance elevates to the level equivalent to English.


Previous researches \cite{dettmers2022gptint, sun2023wanda} have identified two distinctive characteristics within hidden state features of LLMs: first, the presence of large magnitude features, a small set of hidden state features that emerge with significantly larger magnitudes than the remaining ones; and second, the essential nature of these features for the predictive capabilities of LLMs.
Referring to the researches, we analyzed magnitudes of features of MLLMs when inputting few-shot monolingual and translation demonstrations.
Fig. \ref{fig:overlap_layer27} shows the overlap ratios within features between the monolingual and translation demonstrations, and indicates that specific features are predominantly active only when inputting translation demonstrations.
In addition, we will see later that the large magnitude features are relevant for the translation performance.

Motivated by the results, we hypothesized that MLLMs carry out zero-shot inferences while accentuating their alignment capability by forcing them to use large magnitude features that are active when inputting few-shot translation demonstrations.
To achieve this, we retained and pruned weights of MLLMs following \citet{sun2023wanda} that involve operations for the large magnitude features and others, respectively.
We observed that the pruned MLLMs improved zero-shot performance in non-English languages compared to pre-pruned MLLMs. Our contributions in this study are threefold: 
\begin{enumerate}
\item  We demonstrated that specific features exhibit large magnitudes and are predominantly active only when inputting few-shot translation demonstrations. In addition, we showed that the large magnitude features are relevant for performing the translation task.
\item We conducted multilingual zero-shot learning in the Cross-lingual Natural Language Inference (XNLI) and Multilingual Amazon Review Corpus (MARC) tasks while forcing MLLMs to rely on the large magnitude features through pruning. The results indicated that the pruning enhances performance in XGLM and mGPT, but it did not improve the performance of BLOOM.

\item 
Since BLOOM was trained on programming language texts as well as multilingual natural language texts, it has the capability to generate programming language, unlike XGLM and mGPT.
We hypothesized that the capability to generate programming language introduces noise. 
Based on the hypothesis, we attempted to prune weights associated operations for the large magnitude features activated when generating programming language texts.
We observed that it enhances multilingual zero-shot learning performance in BLOOM. 
\end{enumerate}

\section{Task Setting}
\label{section:task_setting}
In this study, we conduct zero-shot learning with MLLMs under \citet{lin-etal-2022-shot}'s scenario.
We consider a language $l \in \mathcal{L}$ and its test example as $x_{l}$.
To perform in-context learning, we convert $x_{l}$ to a cloze-style format that contains a \texttt{[Mask]} symbol using a template $\mathcal{T}$ and map each candidate label $y \in \mathcal{Y}$ into a string using a verbalizer $v: \mathcal{Y} \rightarrow \mathcal{V}^*$.
An input prompt $\mathcal{P} (x_l, y)$ is obtained by substituting the \texttt{[Mask]} symbol in $\mathcal{T} (x_l)$ with $v (y)$.
In zero-shot in-context learning, a prediction $\hat{y}$ is the label with the maximum likelihood:

\begin{equation}
\label{equation:zero_shot}
\hat{y} = \underset{y \in \mathcal{Y}}{\mathrm{argmax}} \, p(\mathcal{P} (x_l, y) \vert \boldsymbol{\theta} )
\end{equation}

\noindent where $\boldsymbol{\theta}$ is parameters of the MLLM. 
The objective of our study is to enhance the non-English language performance of MLLMs in a zero-shot in-context learning setting.


\section{Related Works}


\paragraph{English LLM and their characteristics}
Large language models (LLMs) have shown strong performance in a wide range of downstream tasks. 
While fine-tuning has been a popular approach to adapt models to new tasks, it is often impractical to fine-tune very large models. 
\citet{NEURIPS2020_1457c0d6} proposed zero- and few-shot in-context learning as an alternative, which do not require any gradient updates.
One of the problems with using LLMs is that it requires a lot of computational resources for inference.

One of the problems with using LLMs is that it requires a lot of computational resources for inference. To overcome this problem, several studies have attempted quantization of the LLMs. \citet{dettmers2022gptint} proposed a novel quantization method called \texttt{LLM.int8()} that performs matrix multiplications for large magnitude features and others in 16-bit and 8-bit, respectively. They  empirically demonstrated that \texttt{LLM.int8()} reduces performance degradation compared to performing  all operations in 8-bit. This result suggests that the large magnitude features of LLMs are crucial for their prediction.

Motivated from the success of LLM.int8(), \citet{sun2023wanda} proposed a weights pruning approach named pruning by weights and activations (Wanda). 
Wanda drops weights that do not involve operation for the large magnitude features.
Consider a linear layer's weight of $k$-th layer of a model $\boldsymbol{\theta}^{k} \in \mathbb{R}^{d_\mathrm{out} \times d_\mathrm{in}}$ and hidden states output from previous layer $\boldsymbol{X}^{k - 1} \in \mathbb{R}^{T \times d_{in}}$, where $T$ denotes the number of tokens included in calibration data $X$.
Wanda calculates importance scores $\boldsymbol{S}^{k} \in \mathbb{R}^{d_\mathrm{out} \times d_\mathrm{in}}$ for each element of the weight based on $\boldsymbol{\theta}^{k}$ and $\boldsymbol{X}^{k - 1}$.
Specifically, a score of $i, j$-th element $S_{i,j}$ is calculated as:
\begin{equation}
\label{equation:wanda}
S_{i, j} = \left| \boldsymbol{\theta}^{k}_{i,j} \right| \cdot \lVert \boldsymbol{X}^{k - 1}_j \rVert_2
\end{equation}

\noindent where $\left| \cdot \right|$ represents the absolute value operator, $\lVert \boldsymbol{X}^{k-1}_j \rVert_2$ evaluates the L2-norm of the $j$-th feature vector aggregated across $T$ different tokens.
Wanda prunes weights of the bottom $\alpha$\% scores.
Their experiments demonstrated that Wanda can prune weights of LLMs even mitigating degradation of their performance compared to other pruning methods.

\paragraph{Enhancing multilingual performance of multilingual pre-trained models}
\label{sec:related_word}

As outlined in \S \ref{sec:intro}, several studies have attempted to train Multilingual Large Language Models (MLLMs) using datasets that exhibit a more balanced linguistic distribution compared to English-dominant data used in LLMs. 
Although MLLMs have impressive multilingual capability, their performances in non-English languages do not achieve those in the English level, which is an upper bound level of MLLMs.

This phenomenon has been observed in multilingual pre-trained masked language models, such as mBERT \cite{devlin-etal-2019-bert} and XLM-R \cite{conneau-etal-2020-unsupervised}. Several studies have enhanced the alignment of hidden states between target languages and English using bilingual resources during either the pre-training or fine-tuning phases, demonstrating improved performance in the target languages \cite{lample2019cross,Cao2020Multilingual,yang-etal-2021-bilingual,chi-etal-2021-infoxlm,dou-neubig-2021-word}. Few-shot cross-lingual transfer, which involves fine-tuning the multilingual pre-trained models with a small amount of supervised data in the target language for a downstream task, is a promising approach, and several studies have validated its effectiveness \cite{lauscher-etal-2020-zero,kim-komachi-2023-enhancing}. However, the efficacy of these methodologies when applied to MLLMs remains unverified, and such training approaches require substantial computational resources for adaptation to MLLMs.

\citet{xu-etal-2023-language-representation} undertook research closely aligned with our objective, specifically aiming to enhance non-English performance of MLLMs in a zero-shot learning scenario without the necessity for fine-tuning. 
They introduced a novel methodology termed Language Representation Projection (LRP2). 
LRP2 adjusts the hidden states at the $a$-th layer by subtracting and adding vectors that correspond to the target language and English, respectively. 
Subsequently, the inverse operations are applied at the $b$-th layer.
These vectors are derived from the mean-pooled vectors across all tokens in each respective language’s dataset at the corresponding layers. 

In this research, we provide a promising direction for enhancing non-English performance through comprehensive experiments on various MLLMs.
Specifically, we identify large-magnitude features that are relevant for bringing out the inherent alignment capabilities of MLLMs \cite{lin-etal-2022-shot,etxaniz2023multilingual, vilar-etal-2023-prompting,chitale2024empirical}.
Motivated by the results, we encourage MLLMs to leverage these prominent features by implementing Wanda \cite{sun2023wanda}. Subsequent sections will show that our approach improves the performance of non-English languages across multiple MLLMs, surpassing that of LRP2.

\begin{figure}[t]
 \centering
 \includegraphics[scale=0.4]{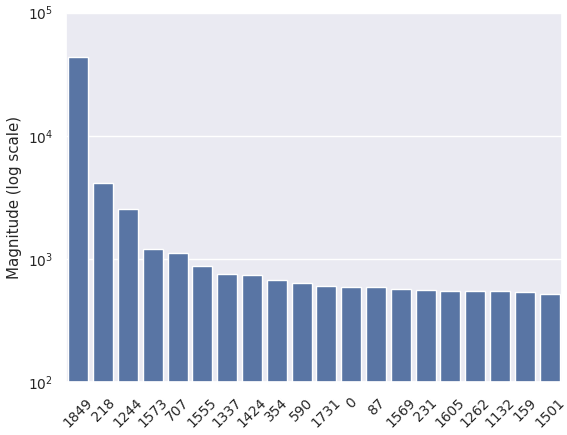}
 \caption{The top 20 dimensions with the largest magnitudes of 27-th layer's features of XGLM activated when inputting $X_\mathrm{Zh-En}$.}\label{fig:mag}
\end{figure}

\section{Detecting Translation Features}
\label{section:detect_feature}
Our challenge is accentuating the alignment capability of MLLMs to enhance their non-English performance even when tasks other than translation.
To explore how to emphasize the alignment capability, we first analyze the behavior of MLLMs during translation.
While previous studies have suggested that large magnitude features are important for inference, they did not reveal whether large magnitude features are the same or different across tasks.
If the large magnitude features that activate exclusively during translation are instrumental for MLLMs' translation performance, the features may be key to bringing out the alignment capability.
Therefore, in this section, we conduct analyses based on the following research questions to show step-by-step that there are large magnitude features that affect for translation performance.

\paragraph{RQ1: Do few-shot translation demonstrations activate specific features?}

Previous studies have suggested that MLLMs can achieve translation performance equivalent to supervised machine translation models by incorporating few-shot translation demonstrations into contextual information.
We consider that salient features when inputting few-shot translation demonstrations play an important role in translation.

Therefore, in this study, we analyze magnitudes of the hidden state features $\boldsymbol{X}^k_{src\text{-}tgt} \in \mathbb{R}^{T_{src\text{-}tgt} \times d_\text{in}}$, those of few-shot translation demonstrations $X_{src\text{-}tgt} = \{x^{1}_{src\text{-}tgt}, ..., x^{N}_{src\text{-}tgt}\}$ output from $k$-th layer of an MLLM.
Here, $T_{src\text{-}tgt}$ is a total amount of tokens included in $X_{src\text{-}tgt}$.
To construct a translation demonstration $x^1_{src\text{-}tgt}$, we use $n$-shot bilingual sentence pairs between $src$ and $tgt$ language randomly sampled from bilingual data.
In addition, we find the hidden state features $\boldsymbol{X}^k_{src} \in \mathbb{R}^{T_{src} \times d_\text{in}}$ and $\boldsymbol{X}^k_{tgt} \in \mathbb{R}^{T_{tgt} \times d_\text{in}}$ , those of few-shot monolingual demonstrations $X_{src} = \{x^1_{src}, ..., x^N_{src}\}$ and $X_{tgt} = \{x^1_{tgt}, ..., x^N_{tgt}\}$, respectively. 
Please refer to  Appendix \ref{appendix:construct_demo} for detailed descriptions on how to construct each demonstration ($x_{src\text{-}tgt}$, $x_{src}$, and $x_{tgt}$).






\begin{figure}[t]
    \centering
    \includegraphics[scale=0.36]{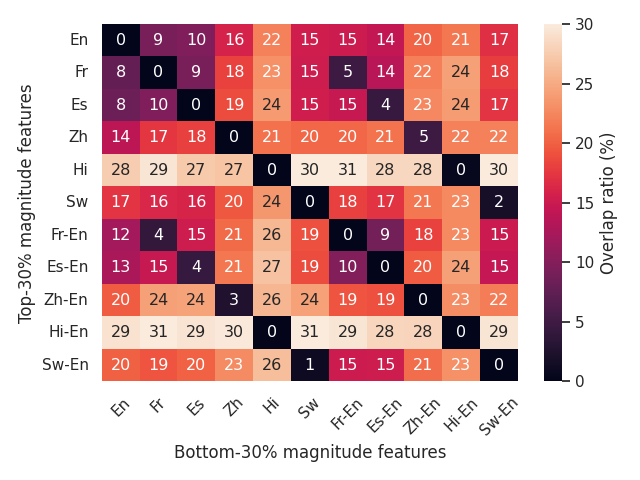}
  \caption{The overlap ratios among the top- and bottom- 30\% features in the 47th layer of XGLM, ranked by magnitude.}\label{fig:overlap_layer47}
\end{figure}

To investigate the magnitude of each feature of $\boldsymbol{X}^k_{src\text{-}tgt}$, $\boldsymbol{X}^k_{src}$, and $\boldsymbol{X}^k_{tgt}$, we find $\lVert \boldsymbol{X}^k_{src\text{-}tgt} \rVert_2, \lVert \boldsymbol{X}^k_{src} \rVert_2, \lVert \boldsymbol{X}^k_{tgt} \rVert_2 \in \mathbb{R}^{d_\text{in}}$ following \citet{sun2023wanda},
which are vectors that each element is L2-norm of the feature aggregated across the tokens of the corresponding hidden states.
Specifically, $j$-th feature of $\lVert \boldsymbol{X}^k_{src\text{-}tgt} \rVert_2$ is $\lVert ( \boldsymbol{X}^{k}_{src\text{-}tgt})_j \rVert_2$.
We examine whether there are features that have extremely large magnitude compared to others.

Furthermore, we measure overlap ratio between the top and bottom $\beta$\% of features, ranked according to their magnitudes. Specifically, the overlap ratio between the features in the top $\beta$\% of $\lVert \boldsymbol{X}^k_{src\text{-}tgt} \rVert_2$ and those in the bottom $\beta$\% of $\lVert \boldsymbol{X}^k_{src} \rVert_2$ is calculated as follows:
%
%
\begin{equation*}
\frac{|\{d^1_{src\text{-}tgt}, \ldots, d^{d_\beta}_{src\text{-}tgt} \} \cap \{d^{d_\mathrm{in} - d_\beta}_{src}, \ldots, d^{d_\mathrm{in}}_{src} \}|}{d_\beta}
\end{equation*}
\noindent where $d_\beta$ ($= d_{\mathrm{in}} \cdot \beta / 100$) is the number of dimensions accounting for $\beta$\% of the total dimensions. The sets $\{d^1_{src\text{-}tgt}, \ldots, d^{d_\beta}_{src\text{-}tgt}\}$ and $ \{d^{d_\mathrm{in} - d_\beta}_{src}, \ldots, d^{d_\mathrm{in}}_{src} \}$ denote dimensions corresponding to the top- and bottom-$d_\beta$ features. 
If the ratio between the top of $\lVert \boldsymbol{X}^k_{src\text{-}tgt} \rVert_2$ and the bottom of $\lVert \boldsymbol{X}^k_{src} \rVert_2$ (or $\lVert \boldsymbol{X}^k_{tgt} \rVert_2$) is high value, it suggests that the features prominently active during translation demonstrations diminish in importance during monolingual demonstrations. 
This implies the existence of large magnitude features that are active only when processing translation demonstrations.

\begin{figure*}[t]
 \centering
 \includegraphics[scale=0.45]{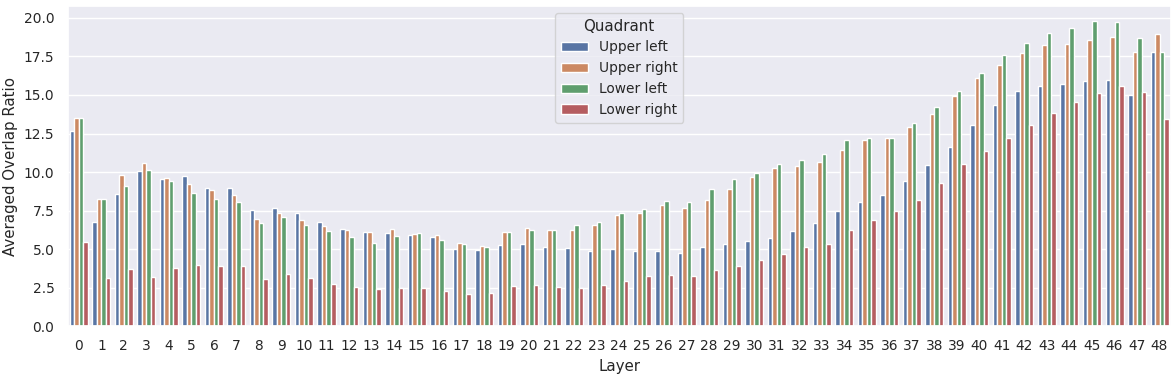}
 \caption{Averaged overlap ratios for each quadrant. This plot quantifies the overlap between monolingual and translation demonstrations in the upper-left, upper-right, lower-left, and lower-right quadrants across different layers. 
  }\label{fig:averaged_overlap}
\end{figure*}

\paragraph{RQ2: Are the large magnitude features relevant for translation performance?}
We investigate whether the large magnitude features when inputting few-shot translation demonstrations are relevant for translation performance.
To reveal this, we retain weights of the MLLM that involve operation for the large magnitude features and prune other weights.
It is accomplished through pruning based on Wanda using $X_{src\text{-}tgt}$ as calibration data.
We denote the pruned weights as $\boldsymbol{\theta}_{src\text{-}tgt}$.
We also perform pruning using $X_{src}$ and $X_{tgt}$ as calibration data, and denote each pruned weights as $\boldsymbol{\theta}_{src}$ and $\boldsymbol{\theta}_{tgt}$, respectively.
We conduct translation with original weights before pruning $\boldsymbol{\theta}$ and each pruned weight.
If $\boldsymbol{\theta}_{src\text{-}tgt}$ maintains the performance relative to $\boldsymbol{\theta}$, and also $\boldsymbol{\theta}_{src}$ and $\boldsymbol{\theta}_{tgt}$ decrease the performance relative to $\boldsymbol{\theta}$, it suggests that it is relevant for translation to use the large magnitude features activated by inputting translation demonstrations.

\subsection{Experimental Settings}
\label{section:detect_feature_setting}
We experimented with XGLM-2.9B, mGPT-1.7B, and BLOOM-3B and employed \citet{etxaniz2023multilingual}'s setting and implementation for the construction of demonstrations and translation.
To construct $X_{src\text{-}tgt}$, $X_{src}$, $X_{tgt}$, we set $N$ and $n$, the numbers of demonstrations and shots of each demonstration, as 100 and 4, respectively.
As a result, we randomly sampled 400 ($= N \times n$) bilingual sentence pairs from development data of FLORES-200 \cite{nllb2022}.

To evaluate translation performance for RQ2, we conducted 4-shot in-context learning.
We constructed a 4-shot translation demonstration $x_{src\text{-}tgt}$ with the procedure described in RQ1.
We incorporated a test example from the source language denoted as $s_\text{test}$ at the end of $x_{src\text{-}tgt}$.
Consequently, a translation was produced by inputting a concatenated string $x_{src\text{-}tgt} \oplus src: s_\text{test} \oplus tgt:$ into each model. 
We used the XNLI test set \cite{conneau-etal-2018-xnli} as evaluation data.
The test sets of each language are bilingual to each other.

For pruning, we used the implementation of \citet{sun2023wanda} and set a pruning ratio $\alpha$ to 30\%. \footnote{See the Appendix \ref{appendix:implementation} for implementation details, including sources of the code and hyperparameters.}

\subsection{Experimental Results}
\paragraph{Answer to RQ1: Few-shot translation demonstrations activate specific features up to middle layer.\footnote{We observed the similar trends as described in this section across other models (i.e. BLOOM and mGPT), layers, and languages.
Please refer Appendix \ref{appendix:translation_feats} and supplemental materials for the remaining results.}}

Fig. \ref{fig:mag} shows that the top-20 dimensions with the largest magnitudes of  $\lVert \boldsymbol{X}^{27}_\mathrm{Zh-En} \rVert_2$.
The figure shows that there are features that have extremely large magnitudes compared to others.
We found that there are common features that have large magnitude independent of the demonstrations, such as 1849, 218, and 1244. 
On the other hand, we observed that there are features included in the top-20 magnitudes of $\lVert \boldsymbol{X}^{27}_\mathrm{Zh-En} \rVert_2$ but not in those of $\lVert \boldsymbol{X}^{27}_\mathrm{Zh} \rVert_2$ and $\lVert \boldsymbol{X}^{27}_\mathrm{En} \rVert_2$, such as 354, 231, and 1262.
We quantified the ratio of unique dimensions within the top-20, top-50, and top-100 magnitudes and showed the results in Appendix \ref{appendix:translation_feats}.

\begin{table*}[tp]
\centering
\small
\tabcolsep 4pt
\scalebox{0.9}{
\begin{tabular}{llllllllllllllll}
\toprule
& \multicolumn{14}{c}{Source Language} & \\
\cmidrule(lr){2-15} 
Weight & Ar & Bg & De & El & Hi & Ru & Sw & Th & Tr & Ur & Vi & Fr & Es & Zh & Avg. \\
\midrule
$\boldsymbol{\theta}$ & 19.4 & 31.3 & 34.3 & 35.4 & 16.7 & 22.5 & 21.4 & 17.3 & 14.8 & 13.2 & 25.0 & 35.2 & 37.3 & 16.1 & 24.3 \\
\cmidrule(lr){1-16} 
$\boldsymbol{\theta}_\mathrm{En}$ & 17.9 & 30.7 & 33.5 & 34.8 & 16.2 & 21.9 & 20.6 & 17.1 & 14.1 & 13.0 & 24.6 & 34.3 & 36.7 & 15.5 & 23.5 \\
$\boldsymbol{\theta}_\mathrm{Fr}$ & 18.0 & 30.2 & 33.3 & 34.4 & 16.1 & 21.8 & 20.4 & 17.1 & 14.2 & 13.1 & 24.7 & 34.1 & 36.2 & 15.0 & 23.5 \\
$\boldsymbol{\theta}_\mathrm{Es}$ & 18.0 & 30.3 & 33.4 & 33.5 & 16.0 & 21.9 & 20.5 & 16.8 & 14.1 & 12.6 & 24.4 & 34.2 & 36.4 & 15.1 & 23.4 \\
$\boldsymbol{\theta}_\mathrm{Zh}$ & 18.0 & 30.2 & 33.5 & 34.3 & 16.0 & 21.9 & 20.2 & 16.9 & 13.9 & 12.8 & 24.4 & 33.8 & 36.2 & 15.2 & 23.4 \\
$\boldsymbol{\theta}_\mathrm{Hi}$ & 18.2 & 30.1 & 33.0 & 33.8 & 15.9 & 22.1 & 20.1 & 17.0 & 14.0 & 12.5 & 23.6 & 33.1 & 35.7 & 15.3 & 23.2 \\
$\boldsymbol{\theta}_\mathrm{Sw}$ & 18.1 & 30.5 & 33.3 & 34.4 & 15.8 & 21.7 & 20.2 & 17.1 & 14.4 & 12.8 & 24.7 & 34.0 & 36.0 & 15.5 & 23.5\\
\cmidrule(lr){1-16} 
$\boldsymbol{\theta}_\mathrm{Fr-En}$ & 18.3$^{\dag\ddag}$ & 31.1$^{\dag\ddag}$ & 33.9$^{\dag\ddag}$ & 35.2$^{\dag\ddag}$ & 16.5$^{\dag\ddag}$ & 22.4$^{\dag\ddag}$ & 20.9$^{\dag\ddag}$ & 17.7$^{\dag\ddag}$ & 14.5$^{\dag\ddag}$ & 13.3$^{\dag}$ & 25.0$^{\dag}$ & 34.6$^{\dag\ddag}$ & 37.0$^{\dag\ddag}$  & 15.7$^{\dag\ddag}$ & 24.1 \\
$\boldsymbol{\theta}_\mathrm{Es-En}$ & 18.6$^{\dag\ddag}$ & 30.9$^{\dag\ddag}$ & 33.9$^{\dag\ddag}$ & 35.0$^{\dag\ddag}$ & 16.5$^{\dag\ddag}$ & 22.3$^{\dag}$ & 20.9$^{\dag\ddag}$ & 17.3$^{\dag\ddag}$ & 14.5$^{\dag}$ & 13.1$^{\ddag}$ & 24.9$^{\dag\ddag}$ & 34.6$^{\dag\ddag}$ & 37.0$^{\dag\ddag}$  & 15.6$^{\dag\ddag}$ & 23.9\\
$\boldsymbol{\theta}_\mathrm{Zh-En}$ & 18.9$^{\dag\ddag}$ & 30.9$^{\dag\ddag}$ & 33.9$^{\dag}$  & 34.9$^{\ddag}$  & 16.5$^{\dag}$ & 22.2$^{\dag}$ & 20.8$^{\ddag}$  & 17.6$^{\dag\ddag}$ & 14.3 & 13.4$^{\dag\ddag}$ & 25.2$^{\dag\ddag}$ & 34.8$^{\dag\ddag}$ & 36.7$^\ddag$ & 15.8$^{\ddag}$ & 24.0 \\
$\boldsymbol{\theta}_\mathrm{Hi-En}$ & 18.7$^{\dag\ddag}$ & 31.1$^{\dag\ddag}$ & 33.8$^{\dag\ddag}$ & 34.9 & 16.3$^{\dag\ddag}$ & 22.2$^{\dag\ddag}$ & 20.7$^{\dag\ddag}$ & 17.5$^{\dag\ddag}$ & 14.4$^{\dag\ddag}$ & 12.9$^{\dag\ddag}$ & 25.1$^{\dag\ddag}$ & 34.4$^{\dag\ddag}$ & 36.9$^{\dag\ddag}$ & 15.7$^{\dag\ddag}$ & 23.9 \\
$\boldsymbol{\theta}_\mathrm{Sw-En}$ & 18.6$^{\dag\ddag}$ & 30.8 & 33.8$^{\dag\ddag}$ & 34.9 & 16.2$^{\dag\ddag}$ & 22.2$^{\dag\ddag}$ & 20.9$^{\dag\ddag}$ & 17.5$^{\dag\ddag}$ & 14.4$^{\dag\ddag}$ & 13.2$^{\dag\ddag}$ & 24.9$^{\dag\ddag}$ & 34.4 & 36.9$^{\dag\ddag}$ & 15.6$^{\dag\ddag}$ & 23.9 \\
\bottomrule
\end{tabular}
}
\caption{BLEU scores of XGLM on original weights $\boldsymbol{\theta}$ and each pruned weights $\boldsymbol{\theta}_{src\text{-}tgt}$, $\boldsymbol{\theta}_{src}$, and $\boldsymbol{\theta}_\mathrm{tgt}$. $\dag$ and $\ddag$ denote statistical significance against $\boldsymbol{\theta}_{src}$ and  $\boldsymbol{\theta}_{tgt}$, respectively. The details for the  statistical significance test were shown in Appendix \ref{appendix:implementation}.
}
\label{table:mt}
\end{table*}

To further investigate the commonalities and differences among features with large magnitudes, we quantified the ratio of overlaps among the top and bottom 30\% of features, ranked by their magnitude, across each demonstration, as depicted in Figures \ref{fig:overlap_layer27} and \ref{fig:overlap_layer47}. 
Up to the middle (27-th) layer, as shown in Fig. \ref{fig:overlap_layer27}, the ratios among the monolingual demonstrations (the upper left quadrant) reveal minimal overlap within linguistically similar languages (En, Fr, and Es), yet exhibit many overlaps within linguistically distant languages, suggesting that the activated features are influenced by linguistic characteristics.
Additionally, two notable observations can be discerned from the figures.
Firstly, the ratios between the monolingual and translation demonstrations (the lower left and upper right quadrants) are relatively high compared to the others, indicating that the features prominently activated by the translation or monolingual demonstrations become less significant in the alternate demonstrations.
Secondly, the ratios within the translation demonstrations (the lower left quadrant) display little overlap, suggesting that similar features are activated by each translation demonstration.
These findings suggest that specific features are uniquely activated during the input of translation demonstrations.

In the proximity of final (47-th) layer, as depicted in Fig. \ref{fig:overlap_layer47}, the ratios were higher across the board.
Interestingly, the ratios between the translation and the source languages' monolingual demonstrations were minimal in these layers.
This result indicates that the features that are active when inputting translation demonstrations are more dependent on non-English languages than on English.

To provide an overarching perspective of the varying patterns across layers, we computed the average overlap ratios for each quadrant of the heatmap, as illustrated in Fig. \ref{fig:averaged_overlap}.
The figure shows that a decline in each overlap ratio from the initial to the middle layers, followed by an increase from the middle to the last layers. 
Moreover, the figure illustrates that the average overlap ratios in the lower quadrant, those among the translation demonstrations, are lower compared to the others across each layer.

\paragraph{Answer to RQ2: The large magnitude features are relevant for maintaining translation performance.}
We conducted to-English translations with the original and pruned models.
Tab. \ref{table:mt} shows the BLEU scores of each model.
The BLEU scores for models pruned by monolingual ($\boldsymbol{\theta}_\mathrm{En}$, $\boldsymbol{\theta}_\mathrm{Fr}$, $\boldsymbol{\theta}_\mathrm{Es}$, etc.) and translation ($\boldsymbol{\theta}_\mathrm{Fr-En}$, $\boldsymbol{\theta}_\mathrm{Es-En}$, etc.) demonstrations were degraded approximately one and 0.3 points compared to the scores of the unpruned original model ($\boldsymbol{\theta}$), respectively.
This result suggests that the features retained in the model pruned based on translation demonstrations, but omitted in the model pruned based on monolingual demonstrations, are important for maintaining the translation performance of the original model.


\section{Multilinguality of Pruned MLLMs}
\label{section:zero-shot}

\begin{table*}[tp]
\centering
\small
\tabcolsep 4pt
\scalebox{0.85}{
\begin{tabular}{lllllllllllllllll}
\toprule
Model & Weight & Ar & Bg & De & El & Hi & Ru & Sw & Th & Tr & Ur & Vi & Fr & Es & Zh & Avg. \\
\midrule
\multirow{9}{*}{XGLM} & $\boldsymbol{\theta}$ & 44.0 & 41.8& 41.6& 44.0 & 44.7& 39.5& 42.5& 44.2& 41.0 & 42.7& 45.2& 45.0 & 35.2& 43.9& 42.5 \\
\cmidrule(lr){2-17} 
& LRP2 & - & - & - & - & 44.6 & - & 42.4 & - & - & - & - &  46.4 & 36.0 & 45.1 & - \\
\cmidrule(lr){2-17} 

& $\boldsymbol{\theta}_\mathrm{Rand}$ & 32.4 & 33.5 & 34.6 & 33.9 & 33.2 & 33.5 & 34.3 & 34 & 33.2 & 34.8 & 33.5 & 36.2 & 34.4 & 33.9 & 33.9 \\
\cmidrule(lr){2-17} 
& $\boldsymbol{\theta}_\mathrm{Fr-En}$ & \textbf{44.9}$^{\dag}$ & 45.8$^{\dag}$ & 42.9$^{\dag}$ & \textbf{46.2}$^{\dag}$ & 44.9 & 43.0$^{\dag}$ & \textbf{43.5}$^{\dag}$ & 45.4$^{\dag}$ & 42.5$^{\dag}$ & 42.8 & \textbf{47.7}$^{\dag}$ & 47.2$^{\dag}$ & 39.7$^{\dag}$ & \textbf{47.2}$^{\dag}$
 & \textbf{44.5} \\
& $\boldsymbol{\theta}_\mathrm{Es-En}$ & 44.3 & \textbf{45.9}$^{\dag}$ & 42.5$^{\dag}$ & 45.6$^{\dag}$ & 44.7& 43.2$^{\dag}$ & 43.3$^{\dag}$ & \textbf{45.9}$^{\dag}$ & \textbf{42.8}$^{\dag}$ & 42.5& 47.5$^{\dag}$ & 47.0$^{\dag}$ & 36.3$^{\dag}$ & 46.8$^{\dag}$ & 44.1 \\
& $\boldsymbol{\theta}_\mathrm{Zh-En}$ & 44.1 & \textbf{45.9}$^{\dag}$ & \textbf{43.0}$^{\dag}$ & 45.9$^{\dag}$ & \textbf{45.0} & \textbf{43.6}$^{\dag}$ & 42.5 & 45.5$^{\dag}$ & 42.2$^{\dag}$ & \textbf{43.0} & 47.5$^{\dag}$ & \textbf{47.7}$^{\dag}$ & \textbf{39.8}$^{\dag}$  & 46.7$^{\dag}$ & 44.4 \\
& $\boldsymbol{\theta}_\mathrm{Hi-En}$ & 43.7 & 42.8$^{\dag}$  & 40.3 & 45.2$^{\dag}$ & 44.7 & 42.5$^{\dag}$ & 42.7 & 45.5$^{\dag}$ & 41.0 & 42.2 & 45.5 & 46.2 & 37.2$^{\dag}$ & 46.0$^{\dag}$  & 43.3 \\
& $\boldsymbol{\theta}_\mathrm{Sw-En}$ & 43.8 & 44.5$^{\dag}$  & 40.3 & 46.1$^{\dag}$ & 43.7 & 41.7$^{\dag}$  & 42.0 & 45.6$^{\dag}$ & 41.7 & 42.0 & 45.4 & 46.4$^{\dag}$ & 38.9$^{\dag}$  & 46.1$^{\dag}$ & 43.4 \\
 \midrule

\multirow{9}{*}{mGPT} & $\boldsymbol{\theta}$ & 39.2 & 39.7 & 35.0 & 41.0 & 38.9 & 39.2 & 34.0 & 41.6 & 39.9 & 39.9 & 42.2 & 42.3 & 39.4 & \textbf{41.8} & 39.6 \\
\cmidrule(lr){2-17} 
& LRP2 & - & - & - & - & 35.2 & - & 34.4 & - & - & - & -& 34.2 & 33.1 & 34.1 & - \\
\cmidrule(lr){2-17} 
& $\boldsymbol{\theta}_\mathrm{Rand}$ & 33.3 & 33.2 & 32.9 & 33.2 & 33.3 & 33.3 & 33.4 & 33.0 & 33.0 & 33.2 & 33.5 & 33.6 & 34.0 & 33.1 & 33.2 \\
\cmidrule(lr){2-17} 
& $\boldsymbol{\theta}_\mathrm{Fr-En}$ & 39.3 & 39.5 & 36.3$^{\dag}$ & 41.9$^{\dag}$ & \textbf{40.2}$^{\dag}$ & 39.5$^{\dag}$ & 34.7$^{\dag}$ & 42.6$^{\dag}$ & \textbf{40.2}$^{\dag}$ & 40.0 & 42.8$^{\dag}$ & 42.1 & 39.7$^{\dag}$ & 41.7 & 40.0 \\
& $\boldsymbol{\theta}_\mathrm{Es-En}$ & \textbf{40.6}$^{\dag}$ & \textbf{40.3}$^{\dag}$ & \textbf{36.6}$^{\dag}$ & \textbf{42.6}$^{\dag}$ & 39.5$^{\dag}$ & \textbf{39.8}$^{\dag}$ & \textbf{35.2}$^{\dag}$ & 42.9$^{\dag}$ & 40.1 & 39.9 & \textbf{43.5}$^{\dag}$ & \textbf{42.5}$^{\dag}$ & \textbf{40.4}$^{\dag}$ & 41.2 & \textbf{40.3} \\
& $\boldsymbol{\theta}_\mathrm{Zh-En}$ & 39.1 & 39.9$^{\dag}$ & 36.1$^{\dag}$ & 41.5$^{\dag}$ & 39.8$^{\dag}$ & 39.0 & 34.4$^{\dag}$ & \textbf{43.4}$^{\dag}$ & 40.1 & 40.2$^{\dag}$ & 43.2$^{\dag}$ & 42.2 & 39.9$^{\dag}$ & 41.4 & 40.0 \\
& $\boldsymbol{\theta}_\mathrm{Hi-En}$  & 39.1 & 39.5 & 34.9 & 41.1 & 40.3$^{\dag}$ & 38.9 & 34.5$^{\dag}$ & 42.1$^{\dag}$ & 40.0 & \textbf{40.5}$^{\dag}$ & 42.6 & 42.1 & 38.9 & 41.6 & 39.7  \\
& $\boldsymbol{\theta}_\mathrm{Sw-En}$ & 38.7 & 39.4 & 34.9 & 40.1 & 40.1$^{\dag}$ & 38.8 & 34.5$^{\dag}$ & 42.3$^{\dag}$ & 39.8 & \textbf{40.7}$^{\dag}$ & 43.6$^{\dag}$ & 42.5$^{\dag}$ & 39.1 & 41.9 & 39.7 \\

 \midrule

\multirow{14}{*}{BLOOM} & $\boldsymbol{\theta}$ & 46.7 & 40.4 & 41.9 & 38.6 & 44.9 & 40.9 & 36.8 & 36.2 & 35.9 & 41.4 & 42.9 & 45.0 & 41.1 & 45.4 & 41.2\\
\cmidrule(lr){2-17} 
& LRP2 & - & - & - & - & 44.6 & - & 37.3 & - & - & - & - & \textbf{46.0} & \textbf{44.3} & \textbf{46.8} & - \\
\cmidrule(lr){2-17} 
& $\boldsymbol{\theta}_\mathrm{Rand}$ & 32.8 & 33.1 & 33.3 & 32.9 & 33.5 & 32.8 & 33.2 & 33.5 & 33.1 & 33.3 & 33.7 & 33.9 & 34.1 & 33.3 & 33.3\\

\cmidrule(lr){2-17} 
& $\boldsymbol{\theta}_\mathrm{Fr-En}$ & 47.0 & 40.3 & 42.2 & 39.5$^{\dag}$ & 45.9$^{\dag}$ & 41.3$^{\dag}$ & 36.9 & 36.5 & 35.6 & 41.1 & 41.9 & 44.9 & 41.2 & 44.6 & 41.3 \\
& $\boldsymbol{\theta}_\mathrm{Es-En}$ & 47.0 & 40.6 & 42.2 & 38.9 & 45.5$^{\dag}$ & 41.1 & 37.1 & 36.6$^{\dag}$ & 35.7 & 40.9 & 41.9 & 44.5 & 41.3 & 44.4 & 41.2 \\
& $\boldsymbol{\theta}_\mathrm{Zh-En}$ & 46.7 & 40.3 & 41.9 & 39.2$^{\dag}$ & 45.3$^{\dag}$ & 41.1 & 37.0 & 36.7$^{\dag}$ & 35.7 & 40.4 & 41.6 & 44.2 & 40.5 & 45.2 & 41.1 \\
& $\boldsymbol{\theta}_\mathrm{Hi-En}$ & 47.2$^{\dag}$ & 40.7 & 40.6 & 40.1$^{\dag}$ & 45.2 & 41.6$^{\dag}$ & 36.0 & \textbf{37.0}$^{\dag}$ & \textbf{36.2} & 41.4 & 42.1 & 45.4$^{\dag}$ & 42.0$^{\dag}$ & 43.7 & 41.4 \\
& $\boldsymbol{\theta}_\mathrm{Sw-En}$ & 46.9 & 40.3 & 40.6 & 40.3$^{\dag}$ & 43.8 & 41.3$^{\dag}$ & 35.7 & 36.2 & 36.1 & 41.5 & 43.0 & 44.4 & 40.8 & 44.1 & 41.0 \\

\cmidrule(lr){2-17} 
& $\boldsymbol{\theta}^\mathrm{Prog}_\mathrm{Fr-En}$ & 46.9 & 40.6 & 42.0 & 40.0$^{\dag}$ & 45.8$^{\dag}$ & \textbf{42.3}$^{\dag}$ & \textbf{37.4}$^{\dag}$ & 36.6$^{\dag}$ & 35.6 & 41.9$^{\dag}$ & 41.3 & 45.1 & 41.6$^{\dag}$ & 45.2 & 41.6 \\
& $\boldsymbol{\theta}^\mathrm{Prog}_\mathrm{Es-En}$ & 47.1$^{\dag}$ & 40.8$^{\dag}$ & 42.2$^{\dag}$ & 39.9$^{\dag}$ & \textbf{46.5}$^{\dag}$ & 41.3$^{\dag}$ & 37.2$^{\dag}$ & 36.3 & 35.5 & 41.2 & 41.1 & 45.1 & 42.2 & 45.6 & 41.6 \\
& $\boldsymbol{\theta}^\mathrm{Prog}_\mathrm{Zh-En}$ & 47.2$^{\dag}$ & 40.5 & 42.2 & 40.0 $^{\dag}$ & 45.8$^{\dag}$ & 41.2 & 37.1 & 36.0 & 35.8 & 41.2 & 42.1 & 45.8$^{\dag}$ & 42.4$^{\dag}$ & 46.3$^{\dag}$ & \textbf{41.7} \\
& $\boldsymbol{\theta}^\mathrm{Prog}_\mathrm{Hi-En}$ & \textbf{47.3}$^{\dag}$ & 40.3 & 41.2 & \textbf{40.3}$^{\dag}$ & 45.4$^{\dag}$ & 41.6$^{\dag}$ & 36.2 & 36.8$^{\dag}$ & 36.1 & 41.3 & 42.2 & 45.2 & 42.1$^{\dag}$ & 45.3 & 41.5 \\
& $\boldsymbol{\theta}^\mathrm{Prog}_\mathrm{Sw-En}$ & 46.8 & \textbf{41.1}$^{\dag}$ & \textbf{42.5}$^{\dag}$ & 39.5$^{\dag}$ & 45.5$^{\dag}$ & 41.2 & 36.9 & 36.6$^{\dag}$ & 35.5 & \textbf{42.5}$^{\dag}$ & \textbf{43.3}$^{\dag}$ & 45.2 & 41.3 & 45.0 & 41.6 \\

\bottomrule
\end{tabular}
}
\caption{Accuracy scores on the XNLI task. The highest scores in each model are indicated in bold.
$\dag$ denotes statistical significance against the original model $\boldsymbol{\theta}$.}
\label{table:zero-shot}
\end{table*}

Our main goal is to enhance the zero-shot performance of MLLMs in non-English contexts by leveraging the alignment capability between English and non-English languages.
In the previous section, we demonstrated that the large magnitude features that are active only when processing few-shot translation demonstrations play an important role to maintain the translation performance of MLLMs.
This implies that these prominent features are essential for bringing out the alignment capability.
We hypothesize that zero-shot in-context learning is performed while accentuating the alignment capability by prioritizing the use of the large magnitude features from few-shot translation demonstrations.
We can accomplish this by employing the pruned weights, denoted as $\boldsymbol{\theta}_{src\text{-}tgt}$, based on few-shot translation demonstrations. 
The pruned model focuses on the large magnitude features relevant for performing translation, disregarding other features during the prediction process, thereby accentuating alignment capability regardless of the target task.
We conduct zero-shot in-context learning based on the pruned weights $\boldsymbol{\theta}_{src\text{-}tgt}$ as follows:

\begin{equation}
\label{equation:zero_shot_pruned_weight}
\hat{y} = \underset{y \in \mathcal{Y}}{\mathrm{argmax}} \, p(\mathcal{P} (x_l, y) | \boldsymbol{\theta}_{src\text{-}tgt} )
\end{equation}


\subsection{Experimental Settings}

We conducted zero-shot in-context learning with the framework as shown in \S \ref{section:task_setting} and followed the \citet{lin-etal-2022-shot}'s setting.
We employed the XNLI \cite{conneau-etal-2018-xnli} and Multilingual Amazon Review Corpus (MARC) \cite{keung-etal-2020-multilingual} tasks to evaluate non-English performance.
For the MARC task, we used the two-label (positive and negative) classification setting.
Please refer to Appendix \ref{appendix:implementation} for the details of a template and verbalizer.

In this experiment, the performances of MLLMs pruned by demonstrations from En, Fr, Es, Zh, Hi, and Sw data were evaluated.
For the construction of demonstrations, MLLMs, and pruning ratio, we used the same setting as described in \S \ref{section:detect_feature_setting}.

For the comparative analysis, we evaluated the performance of randomly pruned models $\boldsymbol{\theta}_\text{Rand}$.
In addition to them, we used LRP2 \cite{xu-etal-2023-language-representation}.
We aligned the data used for applying LRP2 with those used for the pruning process.
Given that LRP2 necessitates data from both source and target languages, we restricted the evaluation to Fr, Es, Zh, Hi, and Sw in the LRP2 experiment.
The hyperparameter ($a$ and $b$) search for LRP2 was conducted using the XNLI development sets corresponding to the respective languages.

\subsection{Experimental Results}
Tables \ref{table:zero-shot} and \ref{table:zero-shot-marc2} show the accuracy scores of each model.
In XGLM and mGPT, the models pruned by the translation demonstrations with high-resource languages ($\boldsymbol{\theta}_\mathrm{Fr-En}$, $\boldsymbol{\theta}_\mathrm{Es-En}$, and $\boldsymbol{\theta}_\mathrm{Zh-En}$) outperformed the original model $\boldsymbol{\theta}$, the randomly pruned models $\boldsymbol{\theta}_\text{Rand}$, and LRP2.\footnote{We observed that the models pruned by the translation demonstrations outperformed those pruned by the monolingual demonstrations. Additionally, we confirmed that this pruning  enhances the performance of a larger-scale model. Please refer Appendix \ref{appendix:multilinguality}  for the detail.}
Additionally, those pruned models showed statistical significance against the original model in more than half of the languages.
While the models pruned by the translation demonstrations with low-resource languages ($\boldsymbol{\theta}_\mathrm{Hi-En}$, $\boldsymbol{\theta}_\mathrm{Sw-En}$) also enhanced the performance, the improvements were marginal.
This result indicates that non-English performances of XGLM and mGPT are effectively enhanced by pruning using translation demonstrations with high-resource languages. 

On the contrary, no discernible enhancement was observed through the application of pruning in the context of BLOOM regardless of the languages.

\begin{table}[tp]
\centering
\small
\tabcolsep 4pt
\scalebox{0.85}{
\begin{tabular}{llllllll}
\toprule
Model & Weight & De & Ja & Fr & Es & Zh & AVg. \\
\midrule
\multirow{8}{*}{XGLM} & $\boldsymbol{\theta}$ & 64.0 & 64.6 & 60.2 & 60.8 & 67.6 & 63.4\\
\cmidrule(lr){2-8} 
& LRP2 & - & - & 60.4 & 60.1 & 66.3 & -  \\
\cmidrule(lr){2-8} 

& $\boldsymbol{\theta}_\text{Rand}$ & 49.3 & 52.3 & 55.4 & 53.4 & 51.1 & 52.3  \\
\cmidrule(lr){2-8} 
& $\boldsymbol{\theta}_\text{Fr-En}$ & 64.4$^{\dag}$ & \textbf{66.9}$^{\dag}$ & \textbf{60.8}$^{\dag}$ & 60.8 & 69.2$^{\dag}$ & \textbf{64.4} \\
& $\boldsymbol{\theta}_\text{Es-En}$ & 64.1 & 66.3$^{\dag}$ & \textbf{60.8}$^{\dag}$ & \textbf{61.0}$^{\dag}$ & 68.7$^{\dag}$ & 64.2 \\
& $\boldsymbol{\theta}_\text{Zh-En}$ & \textbf{64.5}$^{\dag}$ & 64.4 & 60.5$^{\dag}$ & 60.0 & 68.0$^{\dag}$ & 63.5\\
& $\boldsymbol{\theta}_\text{Hi-En}$ & 63.8 & 66.1$^{\dag}$ & 60.0 & 60.4 & \textbf{69.5}$^{\dag}$ & 63.9 \\
& $\boldsymbol{\theta}_\text{Sw-En}$ & 63.9 & 64.8 & 59.8 & 60.3 & 68.5$^{\dag}$ & 63.5 \\
 \midrule

\multirow{8}{*}{mGPT} & $\boldsymbol{\theta}$ & 65.9 & 56.3 & 64.1 & 65.6 & 53.9 & 61.2\\
\cmidrule(lr){2-8} 
& LRP2 & - & - & 51.7 & 52.4 & 50.7 & -  \\
\cmidrule(lr){2-8} 
& $\boldsymbol{\theta}_\text{Rand}$ & 52.8 & 54.4 & 50.3 & 50.8 & 51.9 & 52.0   \\
\cmidrule(lr){2-8} 
& $\boldsymbol{\theta}_\text{Fr-En}$ & \textbf{66.4}$^{\dag}$ & 57.1$^{\dag}$ & 63.6 & 65.7 & 55.2$^{\dag}$ & 61.6 \\
& $\boldsymbol{\theta}_\text{Es-En}$ & \textbf{66.4}$^{\dag}$ & \textbf{57.2}$^{\dag}$ & 64.6$^{\dag}$ & \textbf{65.9}$^{\dag}$ & 55.3$^{\dag}$ & \textbf{61.9} \\
& $\boldsymbol{\theta}_\text{Zh-En}$ & 66.1 & 56.9$^{\dag}$ & \textbf{64.7}$^{\dag}$ & \textbf{65.9}$^{\dag}$ & \textbf{55.5}$^{\dag}$ & 61.8 \\
& $\boldsymbol{\theta}_\text{Hi-En}$ & 66.2$^{\dag}$ & 57.7$^{\dag}$ & 63.7 & 65.6 & 55.4$^{\dag}$ & 61.7 \\
& $\boldsymbol{\theta}_\text{Sw-En}$ & 66.3$^{\dag}$ & 57.4 & 64.2 & 65.8 & 55.2$^{\dag}$ & 61.7 \\
 \midrule

\multirow{13}{*}{BLOOM} & $\boldsymbol{\theta}$ & 52.7 & 59.3 & 62.4 & 63.6 & 63.1 & 60.2 \\
\cmidrule(lr){2-8} 
& LRP2 & - & - & 60.4 & \textbf{63.8} & 66.3 & - \\
\cmidrule(lr){2-8} 
& $\boldsymbol{\theta}_\text{Rand}$ & 51.4 & 53.4 & 52.2 & 52.3 & 52.9 & 52.4 \\
\cmidrule(lr){2-8} 
& $\boldsymbol{\theta}_\text{Fr-En}$ & 52.8 & 59.7$^{\dag}$ & 61.9 & 62.5 & 64.2$^{\dag}$ & 60.3\\
& $\boldsymbol{\theta}_\text{Es-En}$ & \textbf{53.9}$^{\dag}$ & 59.8$^{\dag}$ & 61.3 & 62.0 & 63.6$^{\dag}$ & 60.1 \\
& $\boldsymbol{\theta}_\text{Zh-En}$ & 53.4$^{\dag}$ & 59.4 &  62.1 & 62.7 & 64.6$^{\dag}$ & 60.4 \\
& $\boldsymbol{\theta}_\text{Hi-En}$ & 52.5 & 59.6$^{\dag}$ & 60.9 & 61.6 & 64.6$^{\dag}$ & 59.8 \\
& $\boldsymbol{\theta}_\text{Sw-En}$ & 53.0$^{\dag}$ & 59.6$^{\dag}$ & 61.1 & 62.0 & 63.2 & 59.8 \\

\cmidrule(lr){2-8} 
& $\boldsymbol{\theta}^\text{Prog}_\text{Fr-En}$ & \textbf{53.9}$^{\dag}$ & 59.9$^{\dag}$ & \textbf{62.7}$^{\dag}$ & 63.1 & 63.9$^{\dag}$ & 60.7\\
& $\boldsymbol{\theta}^\text{Prog}_\text{Es-En}$ & 53.6$^{\dag}$ & \textbf{60.3}$^{\dag}$ & 62.5 & 63.2 & 64.9$^{\dag}$ & 60.9 \\
& $\boldsymbol{\theta}^\text{Prog}_\text{Zh-En}$& \textbf{53.9}$^{\dag}$ & 59.9$^{\dag}$ & 62.5 & 63.1 & \textbf{65.6}$^{\dag}$ & \textbf{61.0} \\
& $\boldsymbol{\theta}^\text{Prog}_\text{Hi-En}$& 53.2$^{\dag}$ & 60.2$^{\dag}$ & 61.9 & 62.8 & 64.5$^{\dag}$ & 60.5 \\
& $\boldsymbol{\theta}^\text{Prog}_\text{Sw-En}$& 53.3$^{\dag}$ & 59.6$^{\dag}$ & 61.8 & 62.9 & 63.4$^{\dag}$ & 60.2 \\
\bottomrule
\end{tabular}
}
\caption{Accuracy scores on the MARC task.}
\label{table:zero-shot-marc2}
\end{table}

\subsection{Analysis}
\label{sec:analysis}
\paragraph{Eliminating programming language generation ability.}
While XGLM and mGPT were trained by multilingual natural language texts,  BLOOM was trained by both multilingual natural language and programming language texts.
\citet{bigscience_workshop_2022} showed that BLOOM has the ability to generate programming language text comparable to models such as PolyCoder \cite{10.1145/3520312.3534862}, which is trained using only programming language data.
If the ability for programming language generation persists within the pruned model, it may introduce undesired noise into the model's predictions, such as inference on natural language understanding tasks. 

\begin{table}[tp]
\begin{center}
\scalebox{0.9}{
\centering
\small
\begin{tabular}{lccc}
\toprule
Weight & XGLM & mGPT & BLOOM  \\
\midrule
$\boldsymbol{\theta}$ & 0.741 & 0.537 &  0.561  \\
\midrule
$\boldsymbol{\theta}_\text{Fr-En}$ & 0.744 & 0.593 & 0.569 (0.567)\\
$\boldsymbol{\theta}_\text{Es-En}$ & 0.749 & 0.593 & \textbf{0.576} (0.569)\\
$\boldsymbol{\theta}_\text{Zh-En}$ & \textbf{0.751} & \textbf{0.603} & 0.569 (0.568) \\
\bottomrule
\end{tabular}
}
\end{center}
\caption{Averaged RankC scores with English across each language. 
The BLOOM scores reported both inside and outside of parentheses reflect the results obtained with and without the application of pruning using texts from programming language, respectively.
}
\label{table:averaged_rankc}
\end{table}

In this study, we refine the scoring metric (Eq. \ref{equation:wanda}) of Wanda to reduce the weights that involve operation for such noisy features.
Specifically, we reformulate Eq. \ref{equation:wanda} as follows: 
\begin{equation}
\label{equation:wanda_program}
S_{i, j} = \left| \boldsymbol{\theta}^{k}_{i,j} \right| \cdot \lVert \boldsymbol{X}^{k - 1}_j \rVert_2 \cdot \frac{\lVert \boldsymbol{X}^{k - 1}_j \rVert_2}{\lVert \boldsymbol{Z}^{k - 1}_j \rVert_2 }
\end{equation}
\noindent where $\boldsymbol{Z}^{k - 1}_j \in \mathbb{R}^{T_{Z} \times d_\mathrm{in}}$ represents $(k-1)$-th layer's hidden state features of another calibration data $Z$.
This reformulated equation assigns a small score for $i,j$-th elements if the $j$-th features of $\boldsymbol{Z}^{k - 1}_j$ have large magnitudes.
By selecting calibration data $Z$ that accentuates a model's capacity to perform a specific task, we can eliminate weights associated with operations on features that are activated when executing the task. 
In our scenario, we use programming language texts as $Z$ to eliminate BLOOM's programming language generation capability.

We employed $X_{src\text{-}tgt}$, $X_{src}$, and $X_{tgt}$ as $X$ and Python codes from huggingface as $Z$, and denoted each pruned model as $\boldsymbol{\theta}^\text{Prog}_{src\text{-}tgt}$, $\boldsymbol{\theta}^\text{Prog}_{src}$, and $\boldsymbol{\theta}^\text{Prog}_{tgt}$, respectively.
The models pruned by our reformulated metric demonstrated superior performance compared to those pruned using the original metric proposed by Wanda as shown in Tables \ref{table:zero-shot} and \ref{table:zero-shot-marc2}.
Furthermore, the table reveals that the pruned models, $\boldsymbol{\theta}^\text{Prog}_\text{Fr-En}$, $\boldsymbol{\theta}^\text{Prog}_\text{Es-En}$, and $\boldsymbol{\theta}^\text{Prog}_\text{Zh-En}$, surpassed the performance of $\boldsymbol{\theta}$.
These results indicate that to effectively enhance performance in non-English languages, it is important to selectively retain and prune weights for translation and programming language generation, respectively.
Although the pruned models did not exceed the performance of LRP2, it is noteworthy that the pruning strategy consistently improved non-English performance, unlike LRP2, which significantly degraded the performance in mGPT.
Therefore, our experimental results suggest that pruning is a promising strategy to enhance the non-English performance.

\paragraph{Evaluation of cross-lingual consistency.}
The purpose of pruning MLLMs through translation demonstrations is to elicit alignment ability between English and non-English languages to utilize English inference capability for non-English inference.
Therefore, finally, we measured Ranking based Consistency (RankC) \cite{qi-etal-2023-cross}, which is a metric to measure the consistency of a model's predictions across each language.\footnote{See Appendix \ref{appendix:rankc} for the detailed descriptions of RankC.}
If the RankC scores between English and non-English languages are high, predictions are consistent across the languages, i.e., the model utilizes inference capability in English for non-English inference.

Tab. \ref{table:averaged_rankc} shows the averaged RankC scores with English across each language and employed the XNLI dataset for this experiment.
It illustrates that models pruned using translation demonstrations ($\boldsymbol{\theta}_\mathrm{Fr-En}$, $\boldsymbol{\theta}_\mathrm{Es-En}$, and $\boldsymbol{\theta}_\mathrm{Zh-En}$) achieve superior scores relative to the original model $\boldsymbol{\theta}$. 
This result indicates that the improvement in zero-shot performance for non-English languages stems from more effective utilization of the English inference capabilities than the original model.

\section{Conclusion}
In this study, we showed that there are large magnitude features activated when inputting few-shot translation demonstrations and the pruned MLLMs (i.e. XGLM and mGPT) based on the features enhances zero-shot performance on non-English languages by utilizing the English inference capabilities.
Additionally, we reformulated the scoring metric to eliminate weights associated with operations for large magnitude features in programming language generation and demonstrated that the pruned BLOOM based on the reformulated metric enhances the non-English performance.

The observation from the result of pruning based on the reformulated metric paves the way for further inquiry into the selective pruning of model weights to optimize performance across diverse linguistic tasks.
For future work, we would like to delve deeper into this aspect to identify weights that should be retained or pruned to enhance the performance of MLLMs. 





\section*{Limitations}
\paragraph{Lack of experiments on various hyperparameters and demonstrations.}
In the experiments, the pruning ratio $\alpha$ was fixed to 0.3, and no experiments were conducted with varying ratios. 
By exploring how the performance evolves with different pruning ratios, it's possible to identify the optimal pruning ratio that improves multilingual capabilities. Furthermore, if increasing the pruning ratio enhances performance, it allow for further model size reduction, enabling inference operations with lower memory requirements. 
Therefore, this experiment is crucial for investigating the potential for enhancing the performance of MLLMs and reducing computational costs.
Additionally, we set the parameters $N$ and $n$ to 100 and 4, respectively. 
As the consequence, we used a total of 400 bilingual sentence pairs to construct few-shot translation demonstrations. 
The experiments were conducted under the assumption that the bilingual data were well-prepared, although, in reality, languages with well-prepared bilingual data are scarce. 
If capable of enhancing performance with an even more limited number of bilingual sentence pairs are proved, we can apply the pruning strategy across several language pairs. 
Moreover, since this study did not explor the effect of pruning by varying the values of $N$ and $n$ while keeping the size of the bilingual sentences, the optimal value of each hyperparameter remains unknown.

Previous studies \cite{vilar-etal-2023-prompting,chitale2024empirical} have demonstrated that the quality of few-shot demonstrations significantly impacts the translation performance of multilingual large language models (MLLMs). However, our research did not analyze how changes in the translation demonstrations for the pruning affect the performance. By conducting this analysis, we will provide deeper insights into the sensitivity of MLLMs to variations in translation demonstrations during the pruning process.


\paragraph{Lack of analysis concerning the architectural differences between each model.}
In \S \ref{sec:analysis}, our focus was on the differences in pre-training data. However, an analysis based on architectural differences was not conducted. Notably, BLOOM's architecture is distinctive, especially due to its use of ALiBi, which operates directly on attention scores influenced by token positions. This method of integrating positional information sets BLOOM apart from other models such as XGLM and mGPT. By analyzing these architectural differences, we may reveal the reasons for the distinct trends observed in each model. A detailed analysis focusing on these architectural variances will be undertaken.

\paragraph{Lack of experiments using larger-scale or additionally fine-tuned MLLMs.}
In this study, we focused on evaluating MLLMs with sizes up to 3B parameters. While we observed the effectiveness of the translation demonstration based pruning on XGLM-7.5B, we have not yet evaluated its effectiveness on other larger-scale MLLMs. As a result, it remains uncertain whether the performance improvements observed through the pruning with few-shot translation demonstrations extend to larger models.

In addition, the MLLMs we focused on our study were pre-trained solely on causal language modeling task.
Recent developments have shown that models fine-tuned through instruction tuning \cite{wei2022finetuned} and Reinforcement Learning from Human Feedback (RLHF) \cite{NEURIPS2022_b1efde53} can produce outputs more aligned with human preferences, and it is such models that end users are likely to use. Demonstrating the utility of pruning on these types of models is considered important.

In future studies, we aim to investigate the effectiveness of pruning on larger models and those that have been fine-tuned, to assess its utility further.

\section*{Ethical Considerations}
\paragraph{Potential risks for bias.}
In recent years, several studies have issued warnings about the potential risks associated with pre-trained language models, notably their propensity for generating biased statements. 
Previous researches \cite{zhao-etal-2020-gender, reusens-etal-2023-investigating, goldfarb-tarrant-etal-2023-cross} have shown that biases in multilingual pre-trained models can be transferred across languages. 
Intuitively, enhancing the models' alignment capability between languages, i.e., strengthening the connections between languages, make the transfer of bias more straightforward. 
In this research, we improved the alignment capability between English and non-English languages in MLLMs through pruning, aiming to boost their zero-shot performance in non-English languages. 
Consequently, there is a potential risk that the pruned models might produce statements that include English biases, even when generating content in non-English languages. 
This issue was not considered in our study. 
Therefore, when using the pruned models, sufficient attention should be paid to the problem of bias.

\section*{Acknowledgments}
This work was partly supported by JST, PRESTO Grant Number JPMJPR2366, Japan.

\bibliography{anthology,custom}

\begin{thebibliography}{35}
\expandafter\ifx\csname natexlab\endcsname\relax\def\natexlab#1{#1}\fi

\bibitem[{Ahuja et~al.(2023)Ahuja, Diddee, Hada, Ochieng, Ramesh, Jain, Nambi, Ganu, Segal, Ahmed, Bali, and Sitaram}]{ahuja-etal-2023-mega}
Kabir Ahuja, Harshita Diddee, Rishav Hada, Millicent Ochieng, Krithika Ramesh, Prachi Jain, Akshay Nambi, Tanuja Ganu, Sameer Segal, Mohamed Ahmed, Kalika Bali, and Sunayana Sitaram. 2023.
\newblock \href {https://aclanthology.org/2023.emnlp-main.258} {{MEGA}: Multilingual evaluation of generative {AI}}.
\newblock In \emph{Proceedings of the 2023 Conference on Empirical Methods in Natural Language Processing}, pages 4232--4267, Singapore. Association for Computational Linguistics.

\bibitem[{Ahuja et~al.(2024)Ahuja, Aggarwal, Gumma, Watts, Sathe, Ochieng, Hada, Jain, Ahmed, Bali, and Sitaram}]{ahuja2024megaverse}
Sanchit Ahuja, Divyanshu Aggarwal, Varun Gumma, Ishaan Watts, Ashutosh Sathe, Millicent Ochieng, Rishav Hada, Prachi Jain, Mohamed Ahmed, Kalika Bali, and Sunayana Sitaram. 2024.
\newblock \href {https://doi.org/10.18653/v1/2024.naacl-long.143} {{MEGAVERSE}: Benchmarking large language models across languages, modalities, models and tasks}.
\newblock In \emph{Proceedings of the 2024 Conference of the North American Chapter of the Association for Computational Linguistics: Human Language Technologies (Volume 1: Long Papers)}, pages 2598--2637, Mexico City, Mexico. Association for Computational Linguistics.

\bibitem[{Brown et~al.(2020)Brown, Mann, Ryder, Subbiah, Kaplan, Dhariwal, Neelakantan, Shyam, Sastry, Askell, Agarwal, Herbert-Voss, Krueger, Henighan, Child, Ramesh, Ziegler, Wu, Winter, Hesse, Chen, Sigler, Litwin, Gray, Chess, Clark, Berner, McCandlish, Radford, Sutskever, and Amodei}]{NEURIPS2020_1457c0d6}
Tom Brown, Benjamin Mann, Nick Ryder, Melanie Subbiah, Jared~D Kaplan, Prafulla Dhariwal, Arvind Neelakantan, Pranav Shyam, Girish Sastry, Amanda Askell, Sandhini Agarwal, Ariel Herbert-Voss, Gretchen Krueger, Tom Henighan, Rewon Child, Aditya Ramesh, Daniel Ziegler, Jeffrey Wu, Clemens Winter, Chris Hesse, Mark Chen, Eric Sigler, Mateusz Litwin, Scott Gray, Benjamin Chess, Jack Clark, Christopher Berner, Sam McCandlish, Alec Radford, Ilya Sutskever, and Dario Amodei. 2020.
\newblock \href {https://proceedings.neurips.cc/paper_files/paper/2020/file/1457c0d6bfcb4967418bfb8ac142f64a-Paper.pdf} {Language models are few-shot learners}.
\newblock In \emph{Advances in Neural Information Processing Systems}, volume~33, pages 1877--1901. Curran Associates, Inc.

\bibitem[{Cao et~al.(2020)Cao, Kitaev, and Klein}]{Cao2020Multilingual}
Steven Cao, Nikita Kitaev, and Dan Klein. 2020.
\newblock \href {https://openreview.net/forum?id=r1xCMyBtPS} {Multilingual alignment of contextual word representations}.
\newblock In \emph{International Conference on Learning Representations}.

\bibitem[{Chi et~al.(2021)Chi, Dong, Wei, Yang, Singhal, Wang, Song, Mao, Huang, and Zhou}]{chi-etal-2021-infoxlm}
Zewen Chi, Li~Dong, Furu Wei, Nan Yang, Saksham Singhal, Wenhui Wang, Xia Song, Xian-Ling Mao, Heyan Huang, and Ming Zhou. 2021.
\newblock \href {https://doi.org/10.18653/v1/2021.naacl-main.280} {{I}nfo{XLM}: An information-theoretic framework for cross-lingual language model pre-training}.
\newblock In \emph{Proceedings of the 2021 Conference of the North American Chapter of the Association for Computational Linguistics: Human Language Technologies}, pages 3576--3588, Online. Association for Computational Linguistics.

\bibitem[{Chitale et~al.(2024)Chitale, Gala, and Dabre}]{chitale2024empirical}
Pranjal Chitale, Jay Gala, and Raj Dabre. 2024.
\newblock \href {https://doi.org/10.18653/v1/2024.findings-acl.440} {An empirical study of in-context learning in {LLM}s for machine translation}.
\newblock In \emph{Findings of the Association for Computational Linguistics ACL 2024}, pages 7384--7406, Bangkok, Thailand and virtual meeting. Association for Computational Linguistics.

\bibitem[{Conneau et~al.(2020)Conneau, Khandelwal, Goyal, Chaudhary, Wenzek, Guzm{\'a}n, Grave, Ott, Zettlemoyer, and Stoyanov}]{conneau-etal-2020-unsupervised}
Alexis Conneau, Kartikay Khandelwal, Naman Goyal, Vishrav Chaudhary, Guillaume Wenzek, Francisco Guzm{\'a}n, Edouard Grave, Myle Ott, Luke Zettlemoyer, and Veselin Stoyanov. 2020.
\newblock \href {https://doi.org/10.18653/v1/2020.acl-main.747} {Unsupervised cross-lingual representation learning at scale}.
\newblock In \emph{Proceedings of the 58th Annual Meeting of the Association for Computational Linguistics}, pages 8440--8451, Online. Association for Computational Linguistics.

\bibitem[{Conneau et~al.(2018)Conneau, Rinott, Lample, Williams, Bowman, Schwenk, and Stoyanov}]{conneau-etal-2018-xnli}
Alexis Conneau, Ruty Rinott, Guillaume Lample, Adina Williams, Samuel Bowman, Holger Schwenk, and Veselin Stoyanov. 2018.
\newblock \href {https://doi.org/10.18653/v1/D18-1269} {{XNLI}: Evaluating cross-lingual sentence representations}.
\newblock In \emph{Proceedings of the 2018 Conference on Empirical Methods in Natural Language Processing}, pages 2475--2485, Brussels, Belgium. Association for Computational Linguistics.

\bibitem[{Dettmers et~al.(2022)Dettmers, Lewis, Belkada, and Zettlemoyer}]{dettmers2022gptint}
Tim Dettmers, Mike Lewis, Younes Belkada, and Luke Zettlemoyer. 2022.
\newblock \href {https://openreview.net/forum?id=dXiGWqBoxaD} {{GPT}3.int8(): 8-bit matrix multiplication for transformers at scale}.
\newblock In \emph{Advances in Neural Information Processing Systems}.

\bibitem[{Devlin et~al.(2019)Devlin, Chang, Lee, and Toutanova}]{devlin-etal-2019-bert}
Jacob Devlin, Ming-Wei Chang, Kenton Lee, and Kristina Toutanova. 2019.
\newblock \href {https://doi.org/10.18653/v1/N19-1423} {{BERT}: Pre-training of deep bidirectional transformers for language understanding}.
\newblock In \emph{Proceedings of the 2019 Conference of the North {A}merican Chapter of the Association for Computational Linguistics: Human Language Technologies, Volume 1 (Long and Short Papers)}, pages 4171--4186, Minneapolis, Minnesota. Association for Computational Linguistics.

\bibitem[{Dou and Neubig(2021)}]{dou-neubig-2021-word}
Zi-Yi Dou and Graham Neubig. 2021.
\newblock \href {https://doi.org/10.18653/v1/2021.eacl-main.181} {Word alignment by fine-tuning embeddings on parallel corpora}.
\newblock In \emph{Proceedings of the 16th Conference of the European Chapter of the Association for Computational Linguistics: Main Volume}, pages 2112--2128, Online. Association for Computational Linguistics.

\bibitem[{Enomoto et~al.(2024)Enomoto, Kim, Hirasawa, Nagai, Sato, Nakajima, and Komachi}]{enomoto-etal-2024-tmu}
Taisei Enomoto, Hwichan Kim, Tosho Hirasawa, Yoshinari Nagai, Ayako Sato, Kyotaro Nakajima, and Mamoru Komachi. 2024.
\newblock \href {https://aclanthology.org/2024.bea-1.52} {{TMU}-{HIT} at {MLSP} 2024: How well can {GPT}-4 tackle multilingual lexical simplification?}
\newblock In \emph{Proceedings of the 19th Workshop on Innovative Use of NLP for Building Educational Applications (BEA 2024)}, pages 590--598, Mexico City, Mexico. Association for Computational Linguistics.

\bibitem[{Etxaniz et~al.(2024)Etxaniz, Azkune, Soroa, Lacalle, and Artetxe}]{etxaniz2023multilingual}
Julen Etxaniz, Gorka Azkune, Aitor Soroa, Oier Lacalle, and Mikel Artetxe. 2024.
\newblock \href {https://doi.org/10.18653/v1/2024.naacl-short.46} {Do multilingual language models think better in {E}nglish?}
\newblock In \emph{Proceedings of the 2024 Conference of the North American Chapter of the Association for Computational Linguistics: Human Language Technologies (Volume 2: Short Papers)}, pages 550--564, Mexico City, Mexico. Association for Computational Linguistics.

\bibitem[{Fornaciari et~al.(2022)Fornaciari, Uma, Poesio, and Hovy}]{fornaciari-etal-2022-hard}
Tommaso Fornaciari, Alexandra Uma, Massimo Poesio, and Dirk Hovy. 2022.
\newblock \href {https://doi.org/10.18653/v1/2022.acl-demo.12} {Hard and soft evaluation of {NLP} models with {BOO}t{ST}rap {SA}mpling - {B}oo{S}t{S}a}.
\newblock In \emph{Proceedings of the 60th Annual Meeting of the Association for Computational Linguistics: System Demonstrations}, pages 127--134, Dublin, Ireland. Association for Computational Linguistics.

\bibitem[{Goldfarb-Tarrant et~al.(2023)Goldfarb-Tarrant, Ross, and Lopez}]{goldfarb-tarrant-etal-2023-cross}
Seraphina Goldfarb-Tarrant, Bj{\"o}rn Ross, and Adam Lopez. 2023.
\newblock \href {https://doi.org/10.18653/v1/2023.emnlp-main.346} {Cross-lingual transfer can worsen bias in sentiment analysis}.
\newblock In \emph{Proceedings of the 2023 Conference on Empirical Methods in Natural Language Processing}, pages 5691--5704, Singapore. Association for Computational Linguistics.

\bibitem[{Keung et~al.(2020)Keung, Lu, Szarvas, and Smith}]{keung-etal-2020-multilingual}
Phillip Keung, Yichao Lu, Gy{\"o}rgy Szarvas, and Noah~A. Smith. 2020.
\newblock \href {https://doi.org/10.18653/v1/2020.emnlp-main.369} {The multilingual {A}mazon reviews corpus}.
\newblock In \emph{Proceedings of the 2020 Conference on Empirical Methods in Natural Language Processing (EMNLP)}, pages 4563--4568, Online. Association for Computational Linguistics.

\bibitem[{Kim and Komachi(2023)}]{kim-komachi-2023-enhancing}
Hwichan Kim and Mamoru Komachi. 2023.
\newblock \href {https://doi.org/10.18653/v1/2023.findings-acl.47} {Enhancing few-shot cross-lingual transfer with target language peculiar examples}.
\newblock In \emph{Findings of the Association for Computational Linguistics: ACL 2023}, pages 747--767, Toronto, Canada. Association for Computational Linguistics.

\bibitem[{Koehn(2004)}]{koehn-2004-statistical}
Philipp Koehn. 2004.
\newblock \href {https://aclanthology.org/W04-3250} {Statistical significance tests for machine translation evaluation}.
\newblock In \emph{Proceedings of the 2004 Conference on Empirical Methods in Natural Language Processing}, pages 388--395, Barcelona, Spain. Association for Computational Linguistics.

\bibitem[{Lample and Conneau(2019)}]{lample2019cross}
Guillaume Lample and Alexis Conneau. 2019.
\newblock Cross-lingual language model pretraining.
\newblock \emph{Advances in Neural Information Processing Systems (NeurIPS)}.

\bibitem[{Lauscher et~al.(2020)Lauscher, Ravishankar, Vuli{\'c}, and Glava{\v{s}}}]{lauscher-etal-2020-zero}
Anne Lauscher, Vinit Ravishankar, Ivan Vuli{\'c}, and Goran Glava{\v{s}}. 2020.
\newblock \href {https://doi.org/10.18653/v1/2020.emnlp-main.363} {From zero to hero: {O}n the limitations of zero-shot language transfer with multilingual {T}ransformers}.
\newblock In \emph{Proceedings of the 2020 Conference on Empirical Methods in Natural Language Processing (EMNLP)}, pages 4483--4499, Online. Association for Computational Linguistics.

\bibitem[{Lin et~al.(2022)Lin, Mihaylov, Artetxe, Wang, Chen, Simig, Ott, Goyal, Bhosale, Du, Pasunuru, Shleifer, Koura, Chaudhary, O{'}Horo, Wang, Zettlemoyer, Kozareva, Diab, Stoyanov, and Li}]{lin-etal-2022-shot}
Xi~Victoria Lin, Todor Mihaylov, Mikel Artetxe, Tianlu Wang, Shuohui Chen, Daniel Simig, Myle Ott, Naman Goyal, Shruti Bhosale, Jingfei Du, Ramakanth Pasunuru, Sam Shleifer, Punit~Singh Koura, Vishrav Chaudhary, Brian O{'}Horo, Jeff Wang, Luke Zettlemoyer, Zornitsa Kozareva, Mona Diab, Veselin Stoyanov, and Xian Li. 2022.
\newblock \href {https://doi.org/10.18653/v1/2022.emnlp-main.616} {Few-shot learning with multilingual generative language models}.
\newblock In \emph{Proceedings of the 2022 Conference on Empirical Methods in Natural Language Processing}, pages 9019--9052, Abu Dhabi, United Arab Emirates. Association for Computational Linguistics.

\bibitem[{{NLLB Team} et~al.(2022){NLLB Team}, Costa-jussà, Cross, Çelebi, Elbayad, Heafield, Heffernan, Kalbassi, Lam, Licht, Maillard, Sun, Wang, Wenzek, Youngblood, Akula, Barrault, Mejia-Gonzalez, Hansanti, Hoffman, Jarrett, Sadagopan, Rowe, Spruit, Tran, Andrews, Ayan, Bhosale, Edunov, Fan, Gao, Goswami, Guzmán, Koehn, Mourachko, Ropers, Saleem, Schwenk, and Wang}]{nllb2022}
{NLLB Team}, Marta~R. Costa-jussà, James Cross, Onur Çelebi, Maha Elbayad, Kenneth Heafield, Kevin Heffernan, Elahe Kalbassi, Janice Lam, Daniel Licht, Jean Maillard, Anna Sun, Skyler Wang, Guillaume Wenzek, Al~Youngblood, Bapi Akula, Loic Barrault, Gabriel Mejia-Gonzalez, Prangthip Hansanti, John Hoffman, Semarley Jarrett, Kaushik~Ram Sadagopan, Dirk Rowe, Shannon Spruit, Chau Tran, Pierre Andrews, Necip~Fazil Ayan, Shruti Bhosale, Sergey Edunov, Angela Fan, Cynthia Gao, Vedanuj Goswami, Francisco Guzmán, Philipp Koehn, Alexandre Mourachko, Christophe Ropers, Safiyyah Saleem, Holger Schwenk, and Jeff Wang. 2022.
\newblock \href {https://arxiv.org/abs/2207.04672} {No language left behind: Scaling human-centered machine translation}.
\newblock \emph{arXiv}.

\bibitem[{Ouyang et~al.(2022)Ouyang, Wu, Jiang, Almeida, Wainwright, Mishkin, Zhang, Agarwal, Slama, Ray, Schulman, Hilton, Kelton, Miller, Simens, Askell, Welinder, Christiano, Leike, and Lowe}]{NEURIPS2022_b1efde53}
Long Ouyang, Jeffrey Wu, Xu~Jiang, Diogo Almeida, Carroll Wainwright, Pamela Mishkin, Chong Zhang, Sandhini Agarwal, Katarina Slama, Alex Ray, John Schulman, Jacob Hilton, Fraser Kelton, Luke Miller, Maddie Simens, Amanda Askell, Peter Welinder, Paul~F Christiano, Jan Leike, and Ryan Lowe. 2022.
\newblock \href {https://proceedings.neurips.cc/paper_files/paper/2022/file/b1efde53be364a73914f58805a001731-Paper-Conference.pdf} {Training language models to follow instructions with human feedback}.
\newblock In \emph{Advances in Neural Information Processing Systems}, volume~35, pages 27730--27744. Curran Associates, Inc.

\bibitem[{Papineni et~al.(2002)Papineni, Roukos, Ward, and Zhu}]{papineni-etal-2002-bleu}
Kishore Papineni, Salim Roukos, Todd Ward, and Wei-Jing Zhu. 2002.
\newblock \href {https://doi.org/10.3115/1073083.1073135} {{B}leu: a method for automatic evaluation of machine translation}.
\newblock In \emph{Proceedings of the 40th Annual Meeting of the Association for Computational Linguistics}, pages 311--318, Philadelphia, Pennsylvania, USA. Association for Computational Linguistics.

\bibitem[{Qi et~al.(2023)Qi, Fern{\'a}ndez, and Bisazza}]{qi-etal-2023-cross}
Jirui Qi, Raquel Fern{\'a}ndez, and Arianna Bisazza. 2023.
\newblock \href {https://doi.org/10.18653/v1/2023.emnlp-main.658} {Cross-lingual consistency of factual knowledge in multilingual language models}.
\newblock In \emph{Proceedings of the 2023 Conference on Empirical Methods in Natural Language Processing}, pages 10650--10666, Singapore. Association for Computational Linguistics.

\bibitem[{Reusens et~al.(2023)Reusens, Borchert, Mieskes, De~Weerdt, and Baesens}]{reusens-etal-2023-investigating}
Manon Reusens, Philipp Borchert, Margot Mieskes, Jochen De~Weerdt, and Bart Baesens. 2023.
\newblock \href {https://doi.org/10.18653/v1/2023.emnlp-main.175} {Investigating bias in multilingual language models: Cross-lingual transfer of debiasing techniques}.
\newblock In \emph{Proceedings of the 2023 Conference on Empirical Methods in Natural Language Processing}, pages 2887--2896, Singapore. Association for Computational Linguistics.

\bibitem[{Scao et~al.(2022)Scao, Fan, Akiki, Pavlick, Ilić, Hesslow, Castagné, Luccioni, Yvon, Gallé, Tow, Rush, Biderman, Webson, Ammanamanchi, Wang, Sagot, Muennighoff, del Moral, Ruwase, Bawden, Bekman, McMillan-Major, Beltagy, Nguyen, Saulnier, Tan, Suarez, Sanh, Laurençon, Jernite, Launay, Mitchell, Raffel, Gokaslan, Simhi, Soroa, Aji, Alfassy, Rogers, Nitzav, Xu, Mou, Emezue, Klamm, Leong, van Strien, Adelani, Radev, Ponferrada, Levkovizh, Kim, Natan, De~Toni, Dupont, Kruszewski, Pistilli, Elsahar, Benyamina, Tran, Yu, Abdulmumin, Johnson, Gonzalez-Dios, de~la Rosa, Chim, Dodge, Zhu, Chang, Frohberg, Tobing, Bhattacharjee, Almubarak, Chen, Lo, Von~Werra, Weber, Phan, allal, Tanguy, Dey, Muñoz, Masoud, Grandury, Šaško, Huang, Coavoux, Singh, Jiang, Vu, Jauhar, Ghaleb, Subramani, Kassner, Khamis, Nguyen, Espejel, de~Gibert, Villegas, Henderson, Colombo, Amuok, Lhoest, Harliman, Bommasani, López, Ribeiro, Osei, Pyysalo, Nagel, Bose, Muhammad, Sharma, Longpre, Nikpoor, Silberberg, Pai, Zink,
  Torrent, Schick, Thrush, Danchev, Nikoulina, Laippala, Lepercq, Prabhu, Alyafeai, Talat, Raja, Heinzerling, Si, Salesky, Mielke, Lee, Sharma, Santilli, Chaffin, Stiegler, Datta, Szczechla, Chhablani, Wang, Pandey, Strobelt, Fries, Rozen, Gao, Sutawika, Bari, Al-shaibani, Manica, Nayak, Teehan, Albanie, Shen, Ben-David, Bach, Kim, Bers, Fevry, Neeraj, Thakker, Raunak, Tang, Yong, Sun, Brody, Uri, Tojarieh, Roberts, Chung, Tae, Phang, Press, Li, Narayanan, Bourfoune, Casper, Rasley, Ryabinin, Mishra, Zhang, Shoeybi, Peyrounette, Patry, Tazi, Sanseviero, von Platen, Cornette, Lavallée, Lacroix, Rajbhandari, Gandhi, Smith, Requena, Patil, Dettmers, Baruwa, Singh, Cheveleva, Ligozat, Subramonian, Névéol, Lovering, Garrette, Tunuguntla, Reiter, Taktasheva, Voloshina, Bogdanov, Winata, Schoelkopf, Kalo, Novikova, Forde, Clive, Kasai, Kawamura, Hazan, Carpuat, Clinciu, Kim, Cheng, Serikov, Antverg, van~der Wal, Zhang, Zhang, Gehrmann, Pais, Shavrina, Scialom, Yun, Limisiewicz, Rieser, Protasov, Mikhailov,
  Pruksachatkun, Belinkov, Bamberger, Kasner, Rueda, Pestana, Feizpour, Khan, Faranak, Santos, Hevia, Unldreaj, Aghagol, Abdollahi, Tammour, HajiHosseini, Behroozi, Ajibade, Saxena, Ferrandis, Contractor, Lansky, David, Kiela, Nguyen, Tan, Baylor, Ozoani, Mirza, Ononiwu, Rezanejad, Jones, Bhattacharya, Solaiman, Sedenko, Nejadgholi, Passmore, Seltzer, Sanz, Fort, Dutra, Samagaio, Elbadri, Mieskes, Gerchick, Akinlolu, McKenna, Qiu, Ghauri, Burynok, Abrar, Rajani, Elkott, Fahmy, Samuel, An, Kromann, Hao, Alizadeh, Shubber, Wang, Roy, Viguier, Le, Oyebade, Le, Yang, Nguyen, Kashyap, Palasciano, Callahan, Shukla, Miranda-Escalada, Singh, Beilharz, Wang, Brito, Zhou, Jain, Xu, Fourrier, Periñán, Molano, Yu, Manjavacas, Barth, Fuhrimann, Altay, Bayrak, Burns, Vrabec, Bello, Dash, Kang, Giorgi, Golde, Posada, Sivaraman, Bulchandani, Liu, Shinzato, de~Bykhovetz, Takeuchi, Pàmies, Castillo, Nezhurina, Sänger, Samwald, Cullan, Weinberg, De~Wolf, Mihaljcic, Liu, Freidank, Kang, Seelam, Dahlberg, Broad, Muellner,
  Fung, Haller, Chandrasekhar, Eisenberg, Martin, Canalli, Su, Su, Cahyawijaya, Garda, Deshmukh, Mishra, Kiblawi, Ott, Sang-aroonsiri, Kumar, Schweter, Bharati, Laud, Gigant, Kainuma, Kusa, Labrak, Bajaj, Venkatraman, Xu, Xu, Xu, Tan, Xie, Ye, Bras, Belkada, and Wolf}]{bigscience_workshop_2022}
Teven~Le Scao, Angela Fan, Christopher Akiki, Ellie Pavlick, Suzana Ilić, Daniel Hesslow, Roman Castagné, Alexandra~Sasha Luccioni, François Yvon, Matthias Gallé, Jonathan Tow, Alexander~M. Rush, Stella Biderman, Albert Webson, Pawan~Sasanka Ammanamanchi, Thomas Wang, Benoît Sagot, Niklas Muennighoff, Albert~Villanova del Moral, Olatunji Ruwase, Rachel Bawden, Stas Bekman, Angelina McMillan-Major, Iz~Beltagy, Huu Nguyen, Lucile Saulnier, Samson Tan, Pedro~Ortiz Suarez, Victor Sanh, Hugo Laurençon, Yacine Jernite, Julien Launay, Margaret Mitchell, Colin Raffel, Aaron Gokaslan, Adi Simhi, Aitor Soroa, Alham~Fikri Aji, Amit Alfassy, Anna Rogers, Ariel~Kreisberg Nitzav, Canwen Xu, Chenghao Mou, Chris Emezue, Christopher Klamm, Colin Leong, Daniel van Strien, David~Ifeoluwa Adelani, Dragomir Radev, Eduardo~González Ponferrada, Efrat Levkovizh, Ethan Kim, Eyal~Bar Natan, Francesco De~Toni, Gérard Dupont, Germán Kruszewski, Giada Pistilli, Hady Elsahar, Hamza Benyamina, Hieu Tran, Ian Yu, Idris Abdulmumin,
  Isaac Johnson, Itziar Gonzalez-Dios, Javier de~la Rosa, Jenny Chim, Jesse Dodge, Jian Zhu, Jonathan Chang, Jörg Frohberg, Joseph Tobing, Joydeep Bhattacharjee, Khalid Almubarak, Kimbo Chen, Kyle Lo, Leandro Von~Werra, Leon Weber, Long Phan, Loubna~Ben allal, Ludovic Tanguy, Manan Dey, Manuel~Romero Muñoz, Maraim Masoud, María Grandury, Mario Šaško, Max Huang, Maximin Coavoux, Mayank Singh, Mike Tian-Jian Jiang, Minh~Chien Vu, Mohammad~A. Jauhar, Mustafa Ghaleb, Nishant Subramani, Nora Kassner, Nurulaqilla Khamis, Olivier Nguyen, Omar Espejel, Ona de~Gibert, Paulo Villegas, Peter Henderson, Pierre Colombo, Priscilla Amuok, Quentin Lhoest, Rheza Harliman, Rishi Bommasani, Roberto~Luis López, Rui Ribeiro, Salomey Osei, Sampo Pyysalo, Sebastian Nagel, Shamik Bose, Shamsuddeen~Hassan Muhammad, Shanya Sharma, Shayne Longpre, Somaieh Nikpoor, Stanislav Silberberg, Suhas Pai, Sydney Zink, Tiago~Timponi Torrent, Timo Schick, Tristan Thrush, Valentin Danchev, Vassilina Nikoulina, Veronika Laippala, Violette
  Lepercq, Vrinda Prabhu, Zaid Alyafeai, Zeerak Talat, Arun Raja, Benjamin Heinzerling, Chenglei Si, Elizabeth Salesky, Sabrina~J. Mielke, Wilson~Y. Lee, Abheesht Sharma, Andrea Santilli, Antoine Chaffin, Arnaud Stiegler, Debajyoti Datta, Eliza Szczechla, Gunjan Chhablani, Han Wang, Harshit Pandey, Hendrik Strobelt, Jason~Alan Fries, Jos Rozen, Leo Gao, Lintang Sutawika, M~Saiful Bari, Maged~S. Al-shaibani, Matteo Manica, Nihal Nayak, Ryan Teehan, Samuel Albanie, Sheng Shen, Srulik Ben-David, Stephen~H. Bach, Taewoon Kim, Tali Bers, Thibault Fevry, Trishala Neeraj, Urmish Thakker, Vikas Raunak, Xiangru Tang, Zheng-Xin Yong, Zhiqing Sun, Shaked Brody, Yallow Uri, Hadar Tojarieh, Adam Roberts, Hyung~Won Chung, Jaesung Tae, Jason Phang, Ofir Press, Conglong Li, Deepak Narayanan, Hatim Bourfoune, Jared Casper, Jeff Rasley, Max Ryabinin, Mayank Mishra, Minjia Zhang, Mohammad Shoeybi, Myriam Peyrounette, Nicolas Patry, Nouamane Tazi, Omar Sanseviero, Patrick von Platen, Pierre Cornette, Pierre~François Lavallée,
  Rémi Lacroix, Samyam Rajbhandari, Sanchit Gandhi, Shaden Smith, Stéphane Requena, Suraj Patil, Tim Dettmers, Ahmed Baruwa, Amanpreet Singh, Anastasia Cheveleva, Anne-Laure Ligozat, Arjun Subramonian, Aurélie Névéol, Charles Lovering, Dan Garrette, Deepak Tunuguntla, Ehud Reiter, Ekaterina Taktasheva, Ekaterina Voloshina, Eli Bogdanov, Genta~Indra Winata, Hailey Schoelkopf, Jan-Christoph Kalo, Jekaterina Novikova, Jessica~Zosa Forde, Jordan Clive, Jungo Kasai, Ken Kawamura, Liam Hazan, Marine Carpuat, Miruna Clinciu, Najoung Kim, Newton Cheng, Oleg Serikov, Omer Antverg, Oskar van~der Wal, Rui Zhang, Ruochen Zhang, Sebastian Gehrmann, Shani Pais, Tatiana Shavrina, Thomas Scialom, Tian Yun, Tomasz Limisiewicz, Verena Rieser, Vitaly Protasov, Vladislav Mikhailov, Yada Pruksachatkun, Yonatan Belinkov, Zachary Bamberger, Zdeněk Kasner, Alice Rueda, Amanda Pestana, Amir Feizpour, Ammar Khan, Amy Faranak, Ana Santos, Anthony Hevia, Antigona Unldreaj, Arash Aghagol, Arezoo Abdollahi, Aycha Tammour, Azadeh
  HajiHosseini, Bahareh Behroozi, Benjamin Ajibade, Bharat Saxena, Carlos~Muñoz Ferrandis, Danish Contractor, David Lansky, Davis David, Douwe Kiela, Duong~A. Nguyen, Edward Tan, Emi Baylor, Ezinwanne Ozoani, Fatima Mirza, Frankline Ononiwu, Habib Rezanejad, Hessie Jones, Indrani Bhattacharya, Irene Solaiman, Irina Sedenko, Isar Nejadgholi, Jesse Passmore, Josh Seltzer, Julio~Bonis Sanz, Karen Fort, Livia Dutra, Mairon Samagaio, Maraim Elbadri, Margot Mieskes, Marissa Gerchick, Martha Akinlolu, Michael McKenna, Mike Qiu, Muhammed Ghauri, Mykola Burynok, Nafis Abrar, Nazneen Rajani, Nour Elkott, Nour Fahmy, Olanrewaju Samuel, Ran An, Rasmus Kromann, Ryan Hao, Samira Alizadeh, Sarmad Shubber, Silas Wang, Sourav Roy, Sylvain Viguier, Thanh Le, Tobi Oyebade, Trieu Le, Yoyo Yang, Zach Nguyen, Abhinav~Ramesh Kashyap, Alfredo Palasciano, Alison Callahan, Anima Shukla, Antonio Miranda-Escalada, Ayush Singh, Benjamin Beilharz, Bo~Wang, Caio Brito, Chenxi Zhou, Chirag Jain, Chuxin Xu, Clémentine Fourrier,
  Daniel~León Periñán, Daniel Molano, Dian Yu, Enrique Manjavacas, Fabio Barth, Florian Fuhrimann, Gabriel Altay, Giyaseddin Bayrak, Gully Burns, Helena~U. Vrabec, Imane Bello, Ishani Dash, Jihyun Kang, John Giorgi, Jonas Golde, Jose~David Posada, Karthik~Rangasai Sivaraman, Lokesh Bulchandani, Lu~Liu, Luisa Shinzato, Madeleine~Hahn de~Bykhovetz, Maiko Takeuchi, Marc Pàmies, Maria~A Castillo, Marianna Nezhurina, Mario Sänger, Matthias Samwald, Michael Cullan, Michael Weinberg, Michiel De~Wolf, Mina Mihaljcic, Minna Liu, Moritz Freidank, Myungsun Kang, Natasha Seelam, Nathan Dahlberg, Nicholas~Michio Broad, Nikolaus Muellner, Pascale Fung, Patrick Haller, Ramya Chandrasekhar, Renata Eisenberg, Robert Martin, Rodrigo Canalli, Rosaline Su, Ruisi Su, Samuel Cahyawijaya, Samuele Garda, Shlok~S Deshmukh, Shubhanshu Mishra, Sid Kiblawi, Simon Ott, Sinee Sang-aroonsiri, Srishti Kumar, Stefan Schweter, Sushil Bharati, Tanmay Laud, Théo Gigant, Tomoya Kainuma, Wojciech Kusa, Yanis Labrak, Yash~Shailesh Bajaj,
  Yash Venkatraman, Yifan Xu, Yingxin Xu, Yu~Xu, Zhe Tan, Zhongli Xie, Zifan Ye, Mathilde Bras, Younes Belkada, and Thomas Wolf. 2022.
\newblock \href {https://doi.org/10.48550/ARXIV.2211.05100} {Bloom: A 176b-parameter open-access multilingual language model}.

\bibitem[{Shliazhko et~al.(2024)Shliazhko, Fenogenova, Tikhonova, Kozlova, Mikhailov, and Shavrina}]{10.1162/tacl_a_00633}
Oleh Shliazhko, Alena Fenogenova, Maria Tikhonova, Anastasia Kozlova, Vladislav Mikhailov, and Tatiana Shavrina. 2024.
\newblock \href {https://doi.org/10.1162/tacl_a_00633} {{mGPT: Few-Shot Learners Go Multilingual}}.
\newblock \emph{Transactions of the Association for Computational Linguistics}, 12:58--79.

\bibitem[{Sun et~al.(2024)Sun, Liu, Bair, and Kolter}]{sun2023wanda}
Mingjie Sun, Zhuang Liu, Anna Bair, and J.~Zico Kolter. 2024.
\newblock \href {https://openreview.net/forum?id=PxoFut3dWW} {A simple and effective pruning approach for large language models}.
\newblock In \emph{The Twelfth International Conference on Learning Representations}.

\bibitem[{Vilar et~al.(2023)Vilar, Freitag, Cherry, Luo, Ratnakar, and Foster}]{vilar-etal-2023-prompting}
David Vilar, Markus Freitag, Colin Cherry, Jiaming Luo, Viresh Ratnakar, and George Foster. 2023.
\newblock \href {https://doi.org/10.18653/v1/2023.acl-long.859} {Prompting {P}a{LM} for translation: Assessing strategies and performance}.
\newblock In \emph{Proceedings of the 61st Annual Meeting of the Association for Computational Linguistics (Volume 1: Long Papers)}, pages 15406--15427, Toronto, Canada. Association for Computational Linguistics.

\bibitem[{Wei et~al.(2022)Wei, Bosma, Zhao, Guu, Yu, Lester, Du, Dai, and Le}]{wei2022finetuned}
Jason Wei, Maarten Bosma, Vincent Zhao, Kelvin Guu, Adams~Wei Yu, Brian Lester, Nan Du, Andrew~M. Dai, and Quoc~V Le. 2022.
\newblock \href {https://openreview.net/forum?id=gEZrGCozdqR} {Finetuned language models are zero-shot learners}.
\newblock In \emph{International Conference on Learning Representations}.

\bibitem[{Xu et~al.(2022)Xu, Alon, Neubig, and Hellendoorn}]{10.1145/3520312.3534862}
Frank~F. Xu, Uri Alon, Graham Neubig, and Vincent~Josua Hellendoorn. 2022.
\newblock \href {https://doi.org/10.1145/3520312.3534862} {A systematic evaluation of large language models of code}.
\newblock In \emph{Proceedings of the 6th ACM SIGPLAN International Symposium on Machine Programming}, MAPS 2022, page 1–10, New York, NY, USA. Association for Computing Machinery.

\bibitem[{Xu et~al.(2023)Xu, Li, and Xiong}]{xu-etal-2023-language-representation}
Shaoyang Xu, Junzhuo Li, and Deyi Xiong. 2023.
\newblock \href {https://doi.org/10.18653/v1/2023.emnlp-main.226} {Language representation projection: Can we transfer factual knowledge across languages in multilingual language models?}
\newblock In \emph{Proceedings of the 2023 Conference on Empirical Methods in Natural Language Processing}, pages 3692--3702, Singapore. Association for Computational Linguistics.

\bibitem[{Yang et~al.(2021)Yang, Ma, Cui, Ye, Che, and Wang}]{yang-etal-2021-bilingual}
Ziqing Yang, Wentao Ma, Yiming Cui, Jiani Ye, Wanxiang Che, and Shijin Wang. 2021.
\newblock \href {https://doi.org/10.18653/v1/2021.mrqa-1.10} {Bilingual alignment pre-training for zero-shot cross-lingual transfer}.
\newblock In \emph{Proceedings of the 3rd Workshop on Machine Reading for Question Answering}, pages 100--105, Punta Cana, Dominican Republic. Association for Computational Linguistics.

\bibitem[{Zhao et~al.(2020)Zhao, Mukherjee, Hosseini, Chang, and Hassan~Awadallah}]{zhao-etal-2020-gender}
Jieyu Zhao, Subhabrata Mukherjee, Saghar Hosseini, Kai-Wei Chang, and Ahmed Hassan~Awadallah. 2020.
\newblock \href {https://doi.org/10.18653/v1/2020.acl-main.260} {Gender bias in multilingual embeddings and cross-lingual transfer}.
\newblock In \emph{Proceedings of the 58th Annual Meeting of the Association for Computational Linguistics}, pages 2896--2907, Online. Association for Computational Linguistics.

\end{thebibliography}
\bibliographystyle{acl_natbib}

\appendix

\clearpage

\appendix

\begin{table}[tp]
\begin{center}
\begin{tabular}{ll}
\toprule
Hyperparameters & Value \\
\midrule
Max new tokens & 64 \\
Beam size & 1 \\
Temperature & 0.8 \\
Top $k$ & 100 \\
Top $p$ & 0.75 \\
\bottomrule
\end{tabular}
\end{center}
\caption{Hyperparameters used in translation experiments.}
\label{table:hyperparameter}
\end{table}

\section{How to Construct Few-shot Demonstrations}
\label{appendix:construct_demo}
In this study, we investigated the hidden states from few-shot translation demonstrations $X_{src\text{-}tgt} = \{x^{1}_{src\text{-}tgt}, ..., x^{N}_{src\text{-}tgt}\}$.
Each few-shot demonstration demonstration were constructed  using \citet{etxaniz2023multilingual}'s template function $f_\mathrm{mt}(\cdot)$.
Therefore, a demonstration $x^1_{src\text{-}tgt}$ is as follow:

\begin{eqnarray*}
x^1_{src\text{-}tgt} &=& f_\mathrm{mt}(s^1_1, t^1_1) \oplus ... \oplus f_\text{mt}(s^1_n, t^1_n)  \\
f_\mathrm{mt}(s, t) &=& src : s \oplus  tgt : t \oplus  \text{[EOS]} 
\label{equation:mt_demonstration}
\end{eqnarray*}

\noindent where ``:''  and ``[EOS]'' denote the tokens corresponding to a colon and a end-of-sentence, respectively, and $\oplus$ denotes the concatenation operator. $src$ and $tgt$ represent source and target language names. For example, when English to Chinese translation, $src$ and $tgt$ are ``English'' and ``Chinese''. 

In addition, we examined the the hidden states from few-shot monolingual demonstrations $X_{src} = \{x^1_{src}, ..., x^N_{src}\}$ and $X_{tgt} = \{x^1_{tgt}, ..., x^N_{tgt}\}$.
Here, a demonstration $x^1_{src}$ is as follow:
\begin{eqnarray*}
x^1_{src} &=& f_\mathrm{lm}(s^1_1) \oplus ... \oplus f_\mathrm{lm}(s^1_n)  \\
f_\mathrm{lm}(s) &=& src : s \oplus \text{[EOS]} 
\label{equation:mono_demonstration}
\end{eqnarray*}

\noindent where $f_\mathrm{lm}(\cdot)$ is a template function for the monolingual demonstration.

\section{Implementation Details}
\label{appendix:implementation}
We conducted the experiments with XGLM-2.9B\footnote{\url{https://huggingface.co/facebook/xglm-2.9B}}, mGPT-1.7B\footnote{\url{https://huggingface.co/ai-forever/mGPT}}, and BLOOM-3B\footnote{\url{https://huggingface.co/bigscience/bloom-3b}} from huggingface and used a single Quadro RTX 8000 in the all experiments.

\begin{table}[tp]
\begin{center}
\begin{tabular}{llcc}
\toprule
& & $a$ & $b$ \\
\midrule
\multirow{5}{*}{XGLM} & Fr & 1 & 2 \\
& Es & 8 & 48  \\
& Zh & 1  & 7 \\
& Hi & 1  & 4 \\
& Sw & 7  & 29 \\
\midrule
\multirow{5}{*}{mGPT} & Fr &  0 &  2\\
& Es & 21 & 28  \\
& Zh & 5 & 16 \\
& Hi & 3 & 9 \\
& Sw & 7 & 13 \\
\midrule
\multirow{5}{*}{BLOOM} & Fr & 1  & 22 \\
& Es & 21 & 28  \\
& Zh & 22 & 29 \\
& Hi & 8 & 12 \\
& Sw & 19 & 26 \\
\bottomrule
\end{tabular}
\end{center}
\caption{Optimal layer configurations in LRP2 evaluation. The configurations were searched using the development sets of the corresponding languages.}
\label{table:hyperparameter_lrp2}
\end{table}

\begin{table}[tp]
\centering
\small
\begin{tabular}{lccc}
\toprule
& $\lVert \boldsymbol{X}^{27}_\mathrm{En} \rVert_2$ &  $\lVert \boldsymbol{X}^{27}_\mathrm{Zh} \rVert_2$   \\
\midrule
Top-20 & 30 & 25  \\
Top-50 & 34 & 28  \\
Top-100 & 36 & 25 \\
\bottomrule
\end{tabular}
\caption{The ratio (\%) of unique dimensions within the top-20, top-50, and top-100 magnitudes of $\lVert \boldsymbol{X}^{27}_\mathrm{Zh-En} \rVert_2$ that are absent in the corresponding dimensions of  $\lVert \boldsymbol{X}^{27}_\mathrm{Zh} \rVert_2$ and $\lVert \boldsymbol{X}^{27}_\mathrm{En} \rVert_2$.
}
\label{table:unique_dimension}
\end{table}

To construct each few-shot translation demonstrations and conduct translation experiments, we employed the \citet{etxaniz2023multilingual}'s implementation\footnote{\url{https://github.com/juletx/self-translate}} and the license was not stated.
Tab. \ref{table:hyperparameter} shows the hyperparameters used in our translation experiments and they are the default ones of the implementation.

For pruning using Wanda, we used the official implementation of \citet{sun2023wanda}\footnote{\url{https://github.com/locuslab/wanda}} published by MIT license.
For pruning using refined equation (Eq. \ref{equation:wanda_program}), we used our original implementation and attached the code in supplemental materials.

In our experiments, we used datasets of FLORES-200\footnote{\url{https://huggingface.co/datasets/facebook/flores}}, XNLI\footnote{\url{https://huggingface.co/datasets/xnli}}, MARC\footnote{\url{https://huggingface.co/datasets/SetFit/amazon_reviews_multi_ja}}, and Python codes\footnote{\url{thomwolf/github-python}} from huggingface for constructing of demonstrations, evaluating MLLMs' performance, and measuring the refined importance score, respectively. 

To perform the XNLI task in cloze-style format, we used a template that converts a preliminary $x^\text{pre}$ and a hypothesis $x^\text{hyp}$ to ``$x^\text{pre}, \text{right?}, \text{[Mask]}, x^\text{hyp}$'' and a verbalizer that maps each candidate label (Entailment, Contradiction, and Neutral) to `Yes', `No', and `Also', respectively.
For the MARC task, we used a template that converts a review $x^\text{rev}$  to ``$x^\text{rev} \text{It is } \text{[Mask]}$'' and a verbalizer that maps each candidate label (negative and positive) to `negative', and `positive', respectively.
When selecting a language of template and verbalizer, the language of the test example is expected to be the most intuitive and effective, but previous studies \cite{lin-etal-2022-shot, ahuja-etal-2023-mega, enomoto-etal-2024-tmu} demonstrated that English template and verbalizer achieves the best performance for most test languages.
Therefore, we adopted English template and verbalizer regardless of test languages.

For the evaluation in the translation task, we employed BLEU \cite{papineni-etal-2002-bleu} score and used the huggingface's implementation\footnote{\url{https://huggingface.co/spaces/evaluate-metric/bleu}}.
For the evaluation in the XNLI and MARC tasks, We employed accuracy score and used the scikit-learn implementation\footnote{\url{https://scikit-learn.org/stable/modules/generated/sklearn.metrics.accuracy_score.html}}.
We also performed statistical significance tests through bootstrapping sampling.
We used the \citet{koehn-2004-statistical}\footnote{\url{https://github.com/moses-smt/mosesdecoder/blob/master/scripts/analysis/bootstrap-hypothesis-difference-significance.pl}} and \citet{fornaciari-etal-2022-hard}\footnote{\url{https://github.com/fornaciari/boostsa}}'s implementations for the statistical significance tests for the translation and other tasks, respectively.
In this study, we let that there are statistical significance between the performances when the p-value is less than 0.1.



\section{Detecting Translation Features}
\label{appendix:translation_feats}
\begin{figure*}[t]
  \begin{minipage}[b]{0.33\hsize}
    \centering
    \includegraphics[scale=0.3]{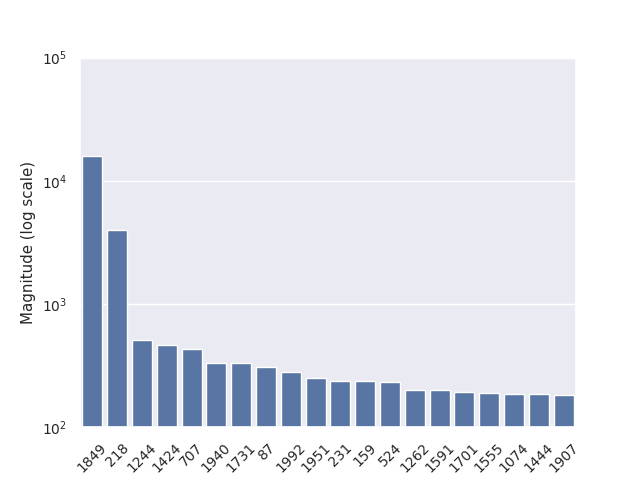}
    \subcaption{$\lVert \boldsymbol{X}^3_\text{Zh} \rVert_2$}
  \end{minipage}
  \begin{minipage}[b]{0.33\hsize}
    \centering
    \includegraphics[scale=0.3]{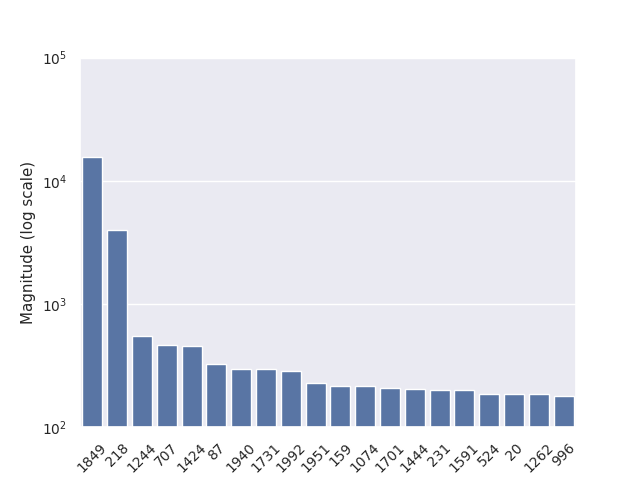}
    \subcaption{$\lVert \boldsymbol{X}^3_\text{En} \rVert_2$}
  \end{minipage}
  \begin{minipage}[b]{0.33\hsize}
    \centering
    \includegraphics[scale=0.3]{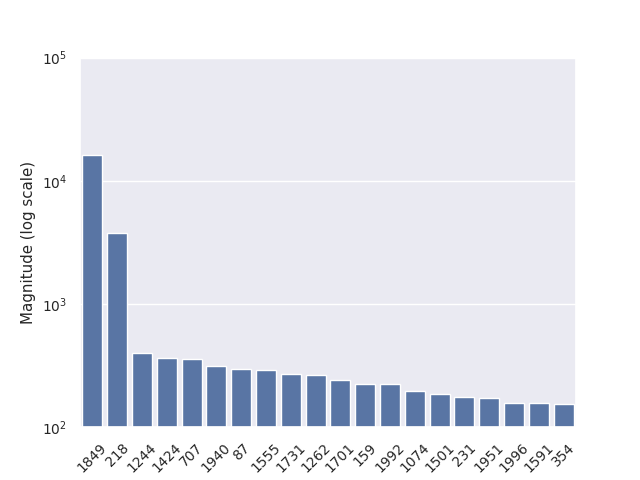}
    \subcaption{$\lVert \boldsymbol{X}^3_\text{Zh-En} \rVert_2$}
  \end{minipage}
  
  \begin{minipage}[b]{0.33\hsize}
    \centering
    \includegraphics[scale=0.3]{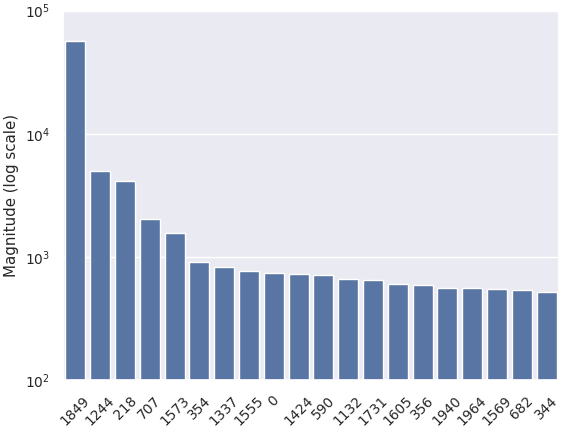}
    \subcaption{$\lVert \boldsymbol{X}^{27}_\text{Zh} \rVert_2$}
  \end{minipage}
  \begin{minipage}[b]{0.33\hsize}
    \centering
    \includegraphics[scale=0.3]{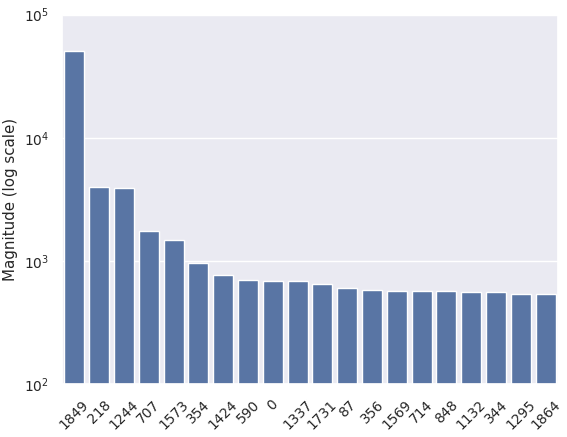}
    \subcaption{$\lVert \boldsymbol{X}^{27}_\text{En} \rVert_2$}
  \end{minipage}
  \begin{minipage}[b]{0.33\hsize}
    \centering
    \includegraphics[scale=0.3]{fig/layer27.xglm.zh-en.png}
    \subcaption{$\lVert \boldsymbol{X}^{27}_\text{Zh-En} \rVert_2$}
  \end{minipage}

  \begin{minipage}[b]{0.33\hsize}
    \centering
    \includegraphics[scale=0.3]{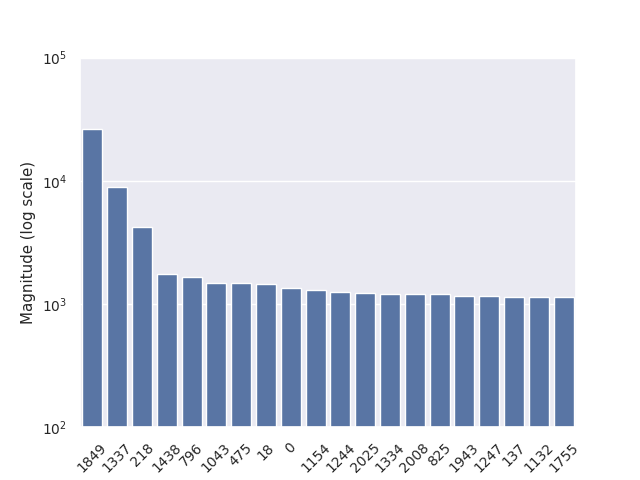}
    \subcaption{$\lVert \boldsymbol{X}^{47}_\text{Zh} \rVert_2$}
  \end{minipage}
  \begin{minipage}[b]{0.33\hsize}
    \centering
    \includegraphics[scale=0.3]{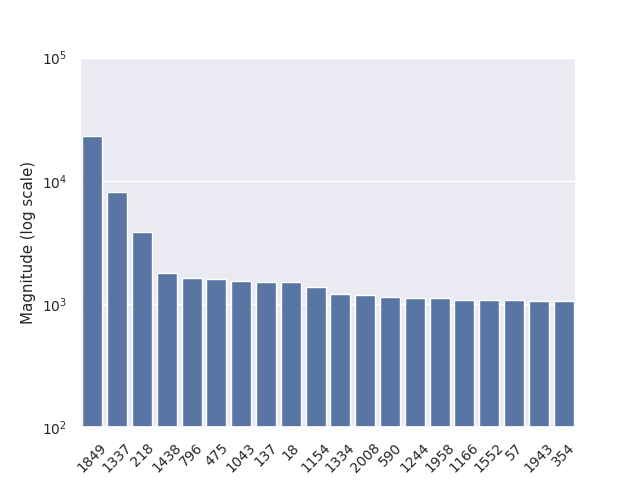}
    \subcaption{$\lVert \boldsymbol{X}^{47}_\text{En} \rVert_2$}
  \end{minipage}
  \begin{minipage}[b]{0.33\hsize}
    \centering
    \includegraphics[scale=0.3]{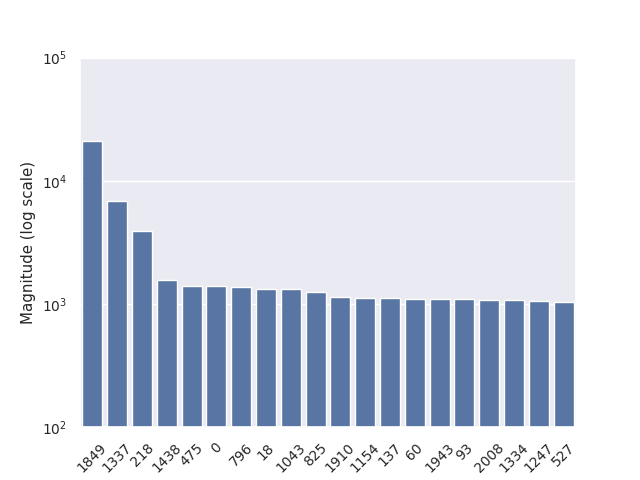}
    \subcaption{$\lVert \boldsymbol{X}^{47}_\text{Zh-En} \rVert_2$}
  \end{minipage}
  
  \caption{The top 20 dimensions of $k$-th layer's features of $\lVert \boldsymbol{X}^{k}_\text{Zh} \rVert_2$, $\lVert \boldsymbol{X}^{k}_\text{En} \rVert_2$ , and $\lVert \boldsymbol{X}^{k}_\text{Zh-En} \rVert_2$ of XGLM.}\label{fig:xglm_mag_zhen}
\end{figure*}

\begin{figure*}[t]
  \begin{minipage}[b]{0.33\hsize}
    \centering
    \includegraphics[scale=0.3]{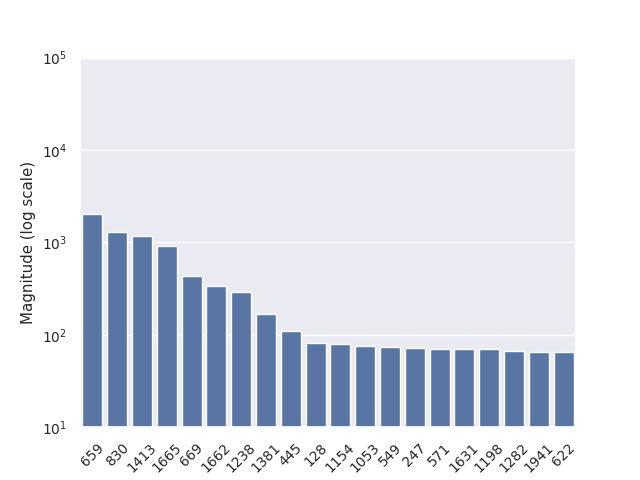}
    \subcaption{$\lVert \boldsymbol{X}^3_\text{Zh} \rVert_2$}
  \end{minipage}
  \begin{minipage}[b]{0.33\hsize}
    \centering
    \includegraphics[scale=0.3]{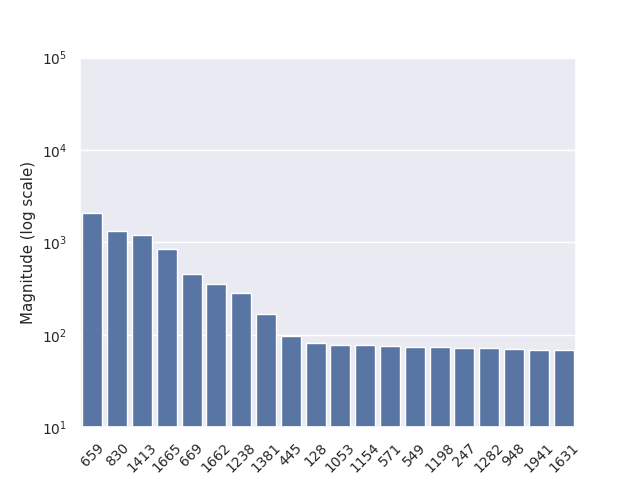}
    \subcaption{$\lVert \boldsymbol{X}^3_\text{En} \rVert_2$}
  \end{minipage}
  \begin{minipage}[b]{0.33\hsize}
    \centering
    \includegraphics[scale=0.3]{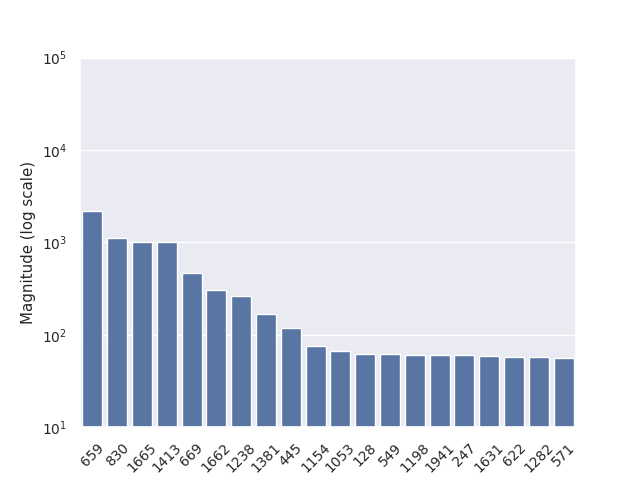}
    \subcaption{$\lVert \boldsymbol{X}^3_\text{Zh-En} \rVert_2$}
  \end{minipage}
  
  \begin{minipage}[b]{0.33\hsize}
    \centering
    \includegraphics[scale=0.3]{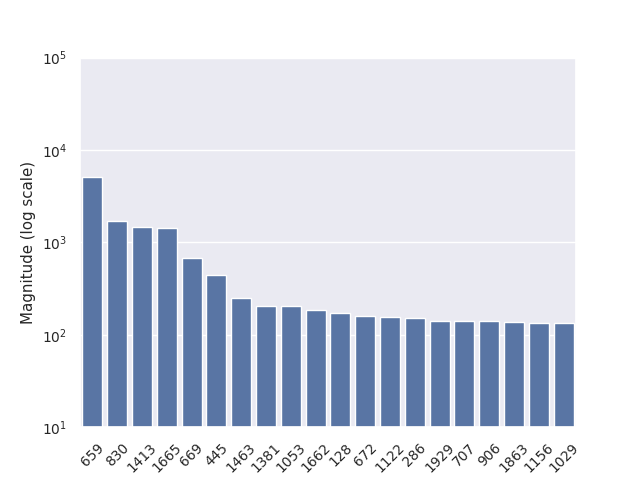}
    \subcaption{$\lVert \boldsymbol{X}^{12}_\text{Zh} \rVert_2$}
  \end{minipage}
  \begin{minipage}[b]{0.33\hsize}
    \centering
    \includegraphics[scale=0.3]{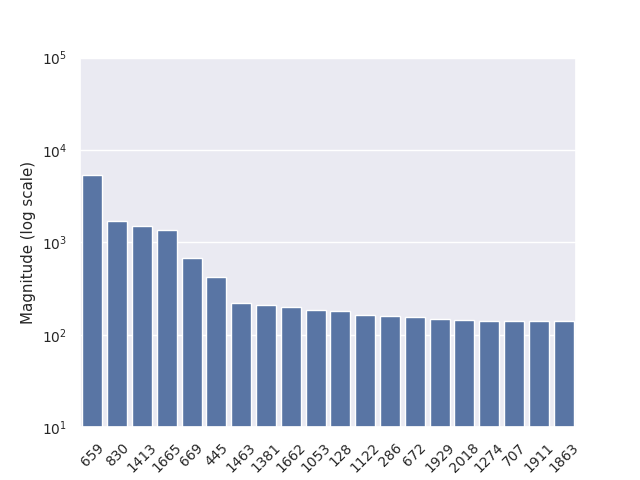}
    \subcaption{$\lVert \boldsymbol{X}^{12}_\text{En} \rVert_2$}
  \end{minipage}
  \begin{minipage}[b]{0.33\hsize}
    \centering
    \includegraphics[scale=0.3]{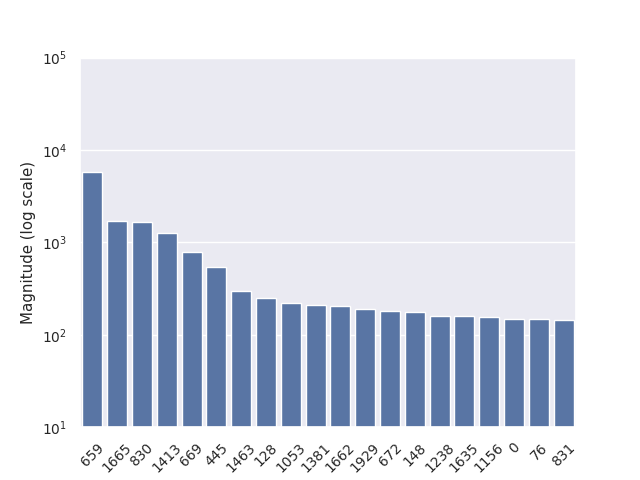}
    \subcaption{$\lVert \boldsymbol{X}^{12}_\text{Zh-En} \rVert_2$}
  \end{minipage}

  \begin{minipage}[b]{0.33\hsize}
    \centering
    \includegraphics[scale=0.3]{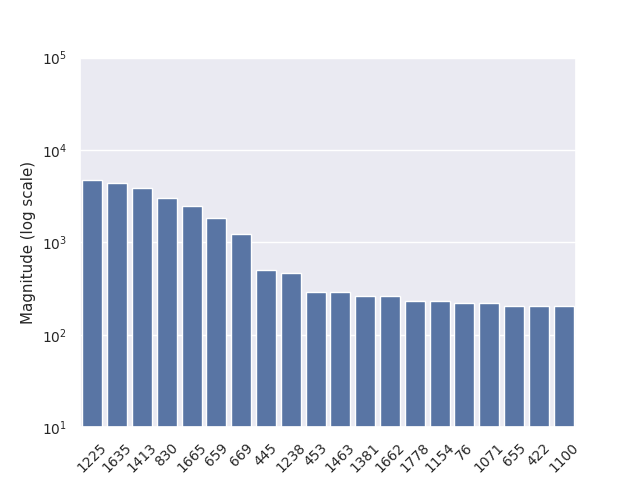}
    \subcaption{$\lVert \boldsymbol{X}^{23}_\text{Zh} \rVert_2$}
  \end{minipage}
  \begin{minipage}[b]{0.33\hsize}
    \centering
    \includegraphics[scale=0.3]{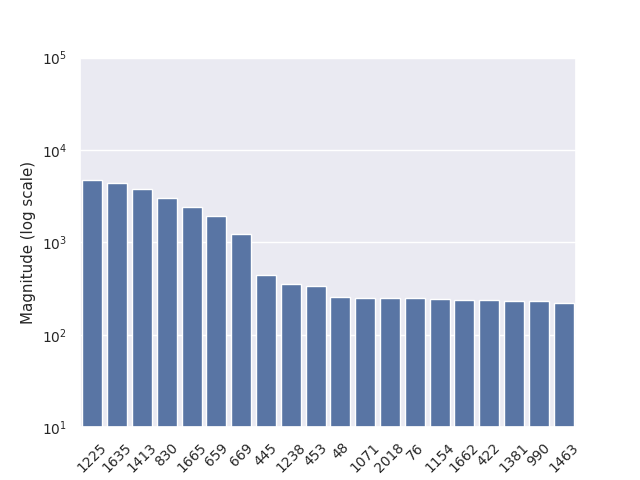}
    \subcaption{$\lVert \boldsymbol{X}^{23}_\text{En} \rVert_2$}
  \end{minipage}
  \begin{minipage}[b]{0.33\hsize}
    \centering
    \includegraphics[scale=0.3]{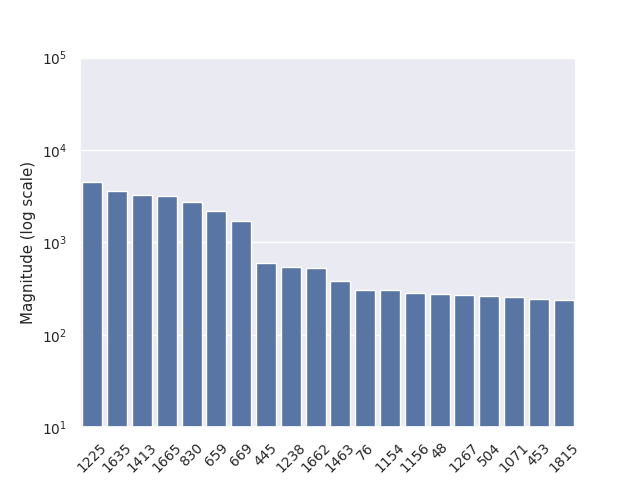}
    \subcaption{$\lVert \boldsymbol{X}^{23}_\text{Zh-En} \rVert_2$}
  \end{minipage}
  
  \caption{The top 20 dimensions of $k$-th layer's features of $\lVert \boldsymbol{X}^{k}_\text{Zh} \rVert_2$, $\lVert \boldsymbol{X}^{k}_\text{En} \rVert_2$ , and $\lVert \boldsymbol{X}^{k}_\text{Zh-En} \rVert_2$ of mGPT.}\label{fig:mgpt_mag_zhen}
\end{figure*}

\begin{figure*}[t]
  \begin{minipage}[b]{0.33\hsize}
    \centering
    \includegraphics[scale=0.3]{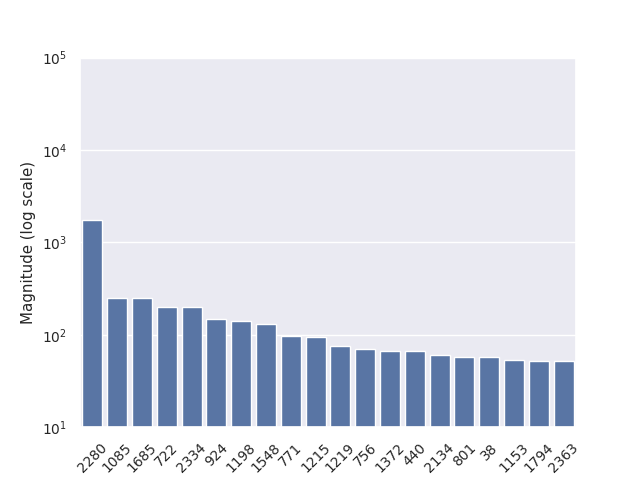}
    \subcaption{$\lVert \boldsymbol{X}^3_\text{Zh} \rVert_2$}
  \end{minipage}
  \begin{minipage}[b]{0.33\hsize}
    \centering
    \includegraphics[scale=0.3]{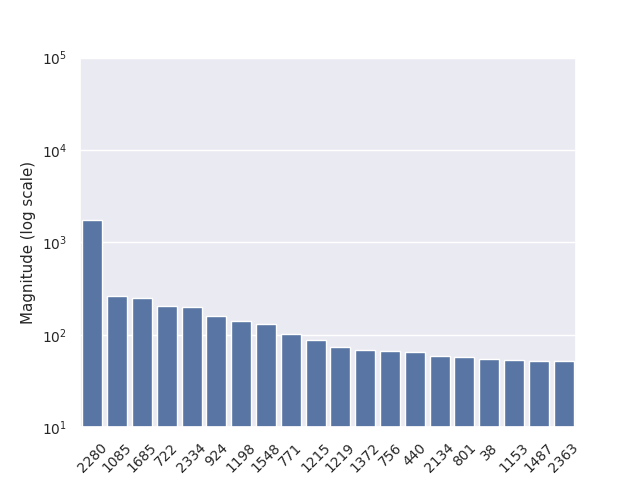}
    \subcaption{$\lVert \boldsymbol{X}^3_\text{En} \rVert_2$}
  \end{minipage}
  \begin{minipage}[b]{0.33\hsize}
    \centering
    \includegraphics[scale=0.3]{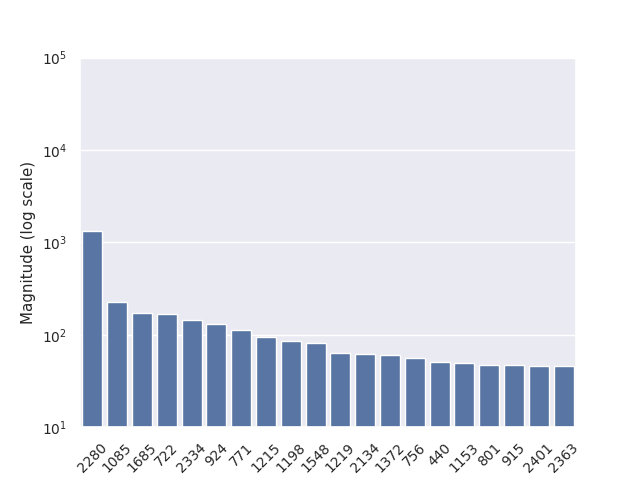}
    \subcaption{$\lVert \boldsymbol{X}^3_\text{Zh-En} \rVert_2$}
  \end{minipage}
  
  \begin{minipage}[b]{0.33\hsize}
    \centering
    \includegraphics[scale=0.3]{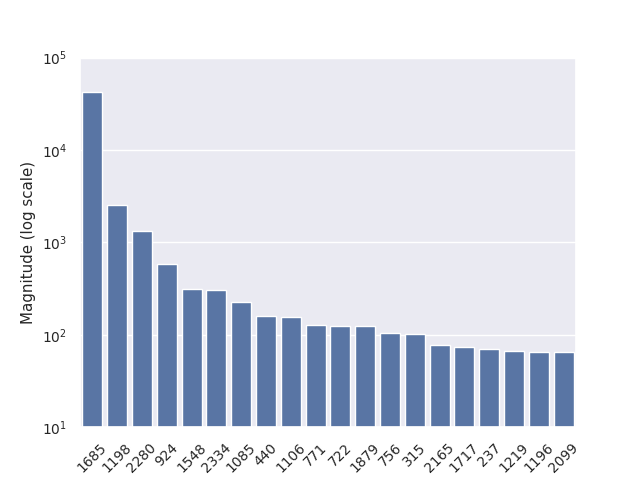}
    \subcaption{$\lVert \boldsymbol{X}^{15}_\text{Zh} \rVert_2$}
  \end{minipage}
  \begin{minipage}[b]{0.33\hsize}
    \centering
    \includegraphics[scale=0.3]{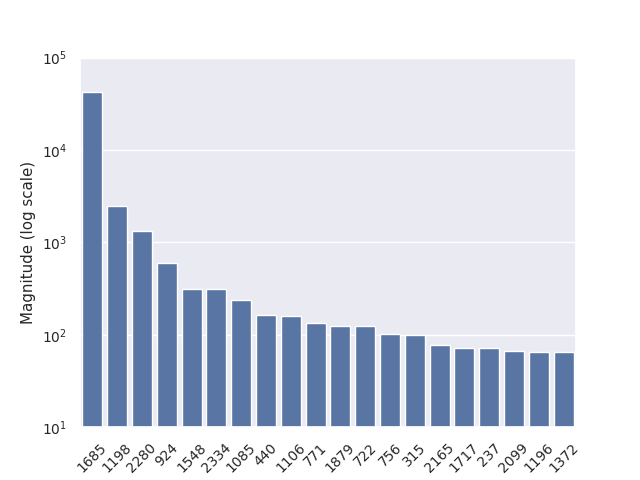}
    \subcaption{$\lVert \boldsymbol{X}^{15}_\text{En} \rVert_2$}
  \end{minipage}
  \begin{minipage}[b]{0.33\hsize}
    \centering
    \includegraphics[scale=0.3]{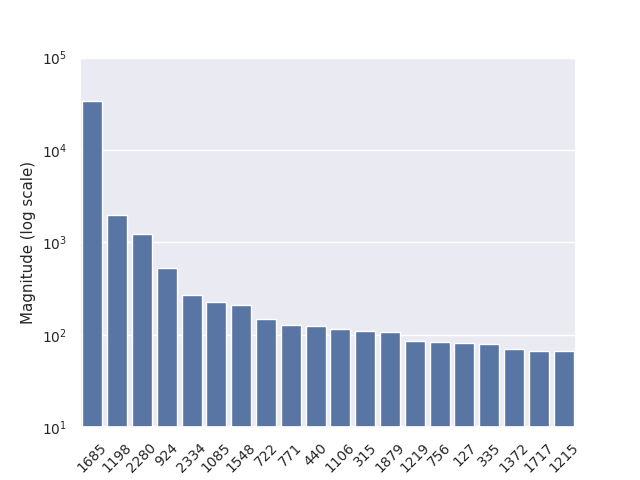}
    \subcaption{$\lVert \boldsymbol{X}^{15}_\text{Zh-En} \rVert_2$}
  \end{minipage}

  \begin{minipage}[b]{0.33\hsize}
    \centering
    \includegraphics[scale=0.3]{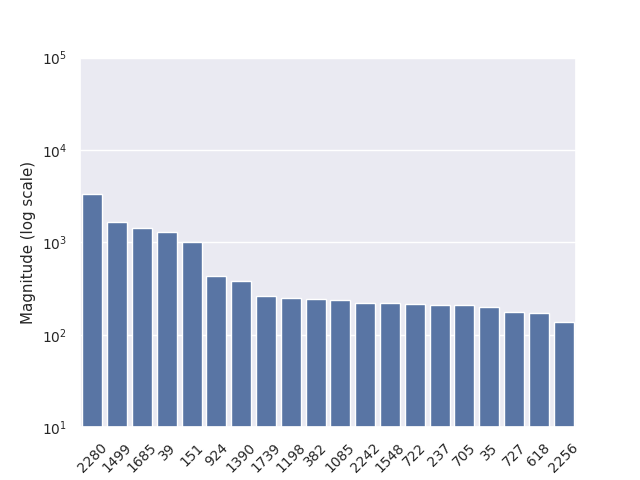}
    \subcaption{$\lVert \boldsymbol{X}^{28}_\text{Zh} \rVert_2$}
  \end{minipage}
  \begin{minipage}[b]{0.33\hsize}
    \centering
    \includegraphics[scale=0.3]{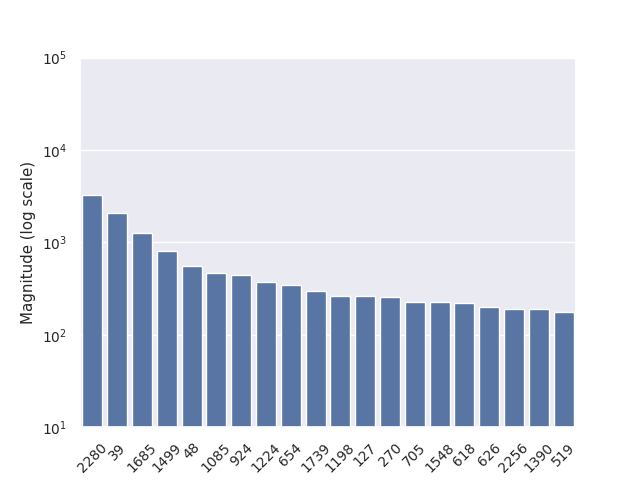}
    \subcaption{$\lVert \boldsymbol{X}^{28}_\text{En} \rVert_2$}
  \end{minipage}
  \begin{minipage}[b]{0.33\hsize}
    \centering
    \includegraphics[scale=0.3]{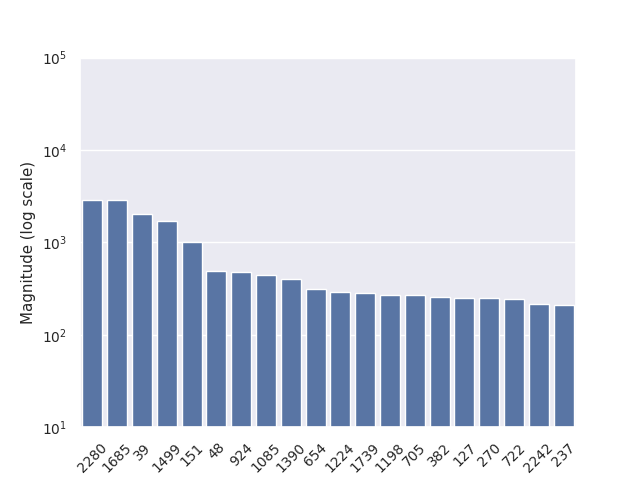}
    \subcaption{$\lVert \boldsymbol{X}^{28}_\text{Zh-En} \rVert_2$}
  \end{minipage}
  
  \caption{The top 20 dimensions of $k$-th layer's features of $\lVert \boldsymbol{X}^{k}_\text{Zh} \rVert_2$, $\lVert \boldsymbol{X}^{k}_\text{En} \rVert_2$ , and $\lVert \boldsymbol{X}^{k}_\text{Zh-En} \rVert_2$ of BLOOM.}\label{fig:bloom_mag_zhen}
\end{figure*}

Figures \ref{fig:xglm_mag_zhen}, \ref{fig:mgpt_mag_zhen}, and \ref{fig:bloom_mag_zhen} show the top-20 dimensions with the largest magnitudes of $k$-th layer's features of $\lVert \boldsymbol{X}^{k}_\text{Zh} \rVert_2$, $\lVert \boldsymbol{X}^{k}_\text{En} \rVert_2$ , and $\lVert \boldsymbol{X}^{k}_\text{Zh-En} \rVert_2$.
Each figure corresponds to XGLM, mGPT, and BLOOM, respectively.
The figures demonstrate that there are features that have extremely large magnitudes compared to others regardless of each MLLM.

\begin{figure*}[t]
  \begin{minipage}[b]{0.33\hsize}
    \centering
    \includegraphics[scale=0.33]{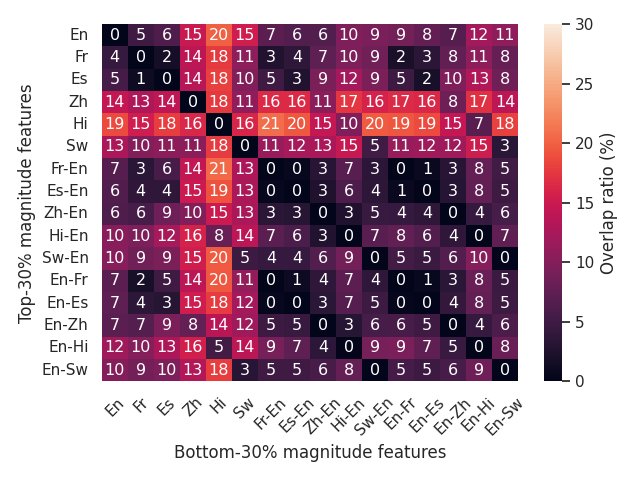}
    \subcaption{Third layer}
  \end{minipage}
  \begin{minipage}[b]{0.33\hsize}
    \centering
    \includegraphics[scale=0.33]{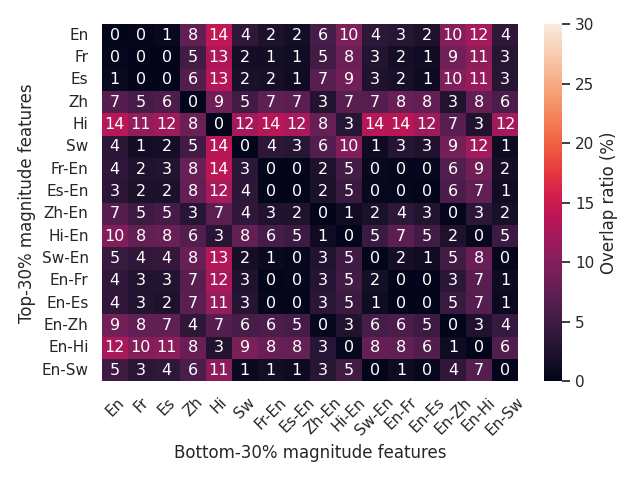}
    \subcaption{20-th layer}
  \end{minipage}
  \begin{minipage}[b]{0.33\hsize}
    \centering
    \includegraphics[scale=0.33]{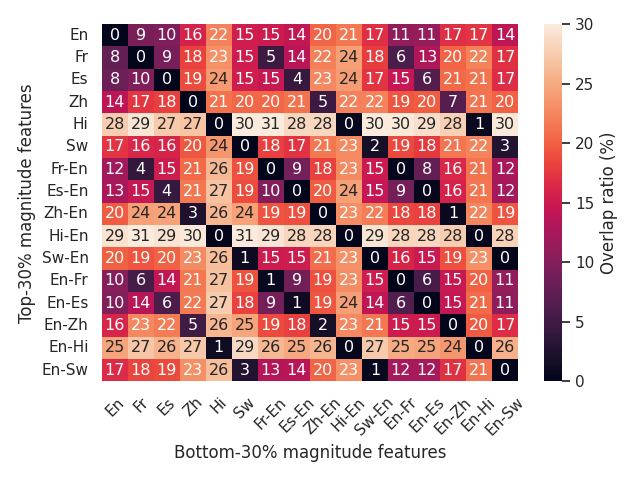}
    \subcaption{47-th layer}
  \end{minipage}
  \caption{The overlap ratios among the top- and bottom-30\% features of XGLM ranked by magnitude. 
  }\label{fig:overlap_xglm_all}
\end{figure*}

\begin{figure*}[t]
  \begin{minipage}[b]{0.33\hsize}
    \centering
    \includegraphics[scale=0.3]{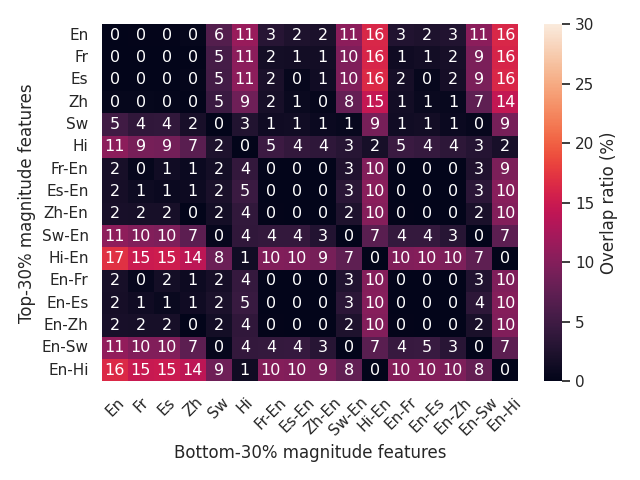}
    \subcaption{Third layer}
  \end{minipage}
  \begin{minipage}[b]{0.33\hsize}
    \centering
    \includegraphics[scale=0.3]{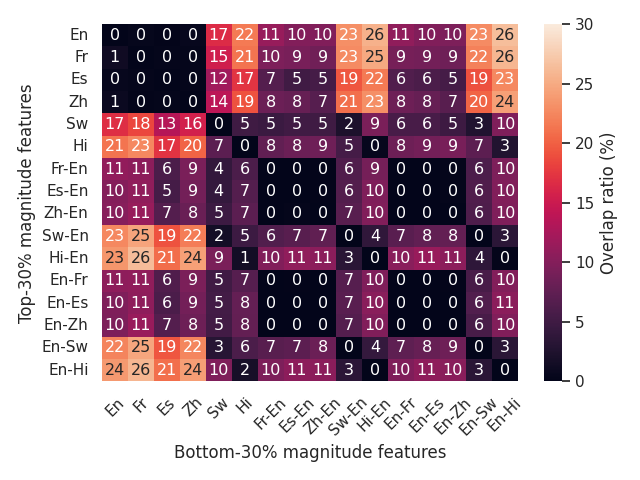}
    \subcaption{12-th layer}
  \end{minipage}
  \begin{minipage}[b]{0.33\hsize}
    \centering
    \includegraphics[scale=0.3]{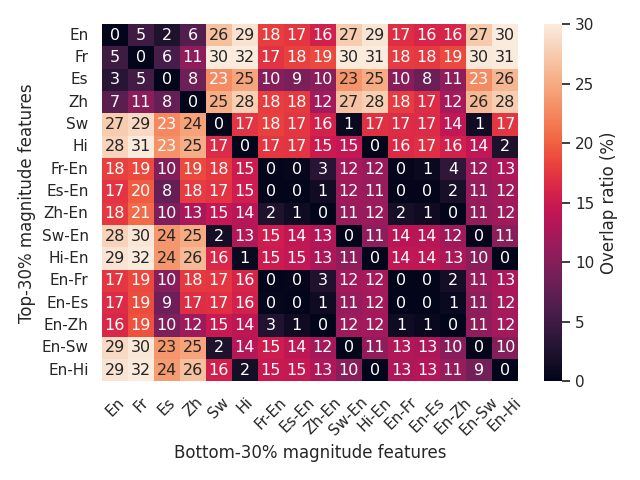}
    \subcaption{23-th layer}
  \end{minipage}
  \caption{The overlap ratios within the top- and bottom-30\% features of mGPT ranked by magnitude.}\label{fig:mgpt_overlap}
\end{figure*}

\begin{figure*}[t]
  \begin{minipage}[b]{0.33\hsize}
    \centering
    \includegraphics[scale=0.3]{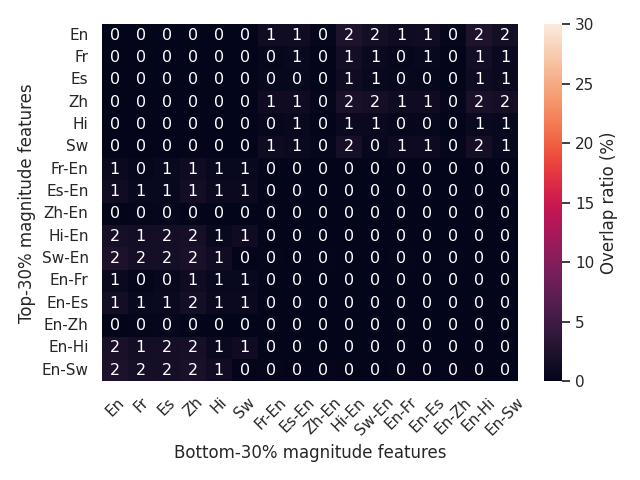}
    \subcaption{Third layer}
  \end{minipage}
  \begin{minipage}[b]{0.33\hsize}
    \centering
    \includegraphics[scale=0.3]{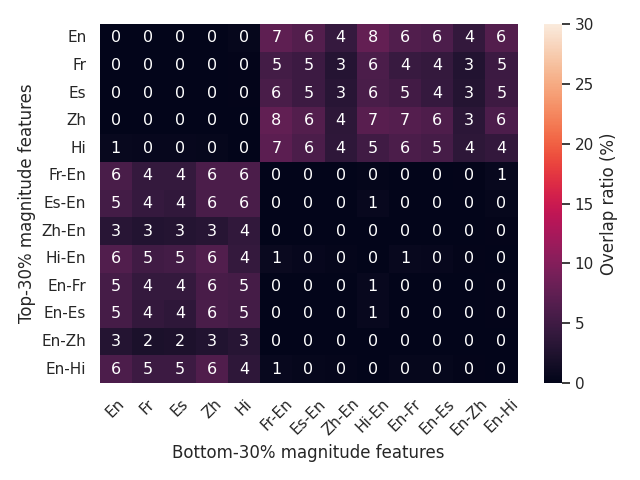}
    \subcaption{20-th layer}
  \end{minipage}
  \begin{minipage}[b]{0.33\hsize}
    \centering
    \includegraphics[scale=0.3]{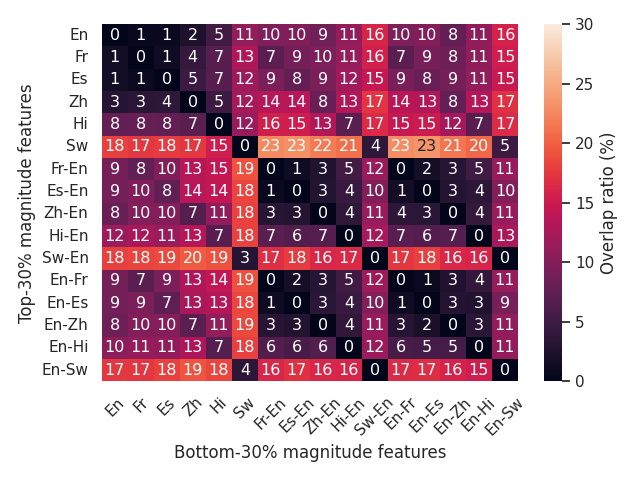}
    \subcaption{28-th layer}
  \end{minipage}
  \caption{The overlap ratios within the top- and bottom-30\% features of BLOOM ranked by magnitude. }\label{fig:bloom_overlap}
\end{figure*}

Figures \ref{fig:overlap_xglm_all}, \ref{fig:mgpt_overlap} and \ref{fig:bloom_overlap} are heatmaps illustrated the overlap ratios within top- and bottom 30\% of features $\lVert \boldsymbol{X}^{k}_{src} \rVert_2$, $\lVert \boldsymbol{X}^{k}_{tgt} \rVert_2$ , and $\lVert \boldsymbol{X}^{k}_{src\text{-}tgt} \rVert_2$ ranked by magnitude.
Each figure corresponds to XGLM, mGPT and BLOOM, respectively.
In XGLM and BLOOM, the ratios indicate that the overlaps within the monolingual demonstrations (the upper left quadrant) and translation demonstrations (the lower right quadrant) are smaller compared to those observed between the monolingual and translation demonstrations (the lower left and upper right quadrants). In mGPT, while the ratios associated with English, French, Spanish, and Chinese languages exhibit similar trends to those in XGLM and BLOOM, the ratios linked with Hindi and Swahili demonstrate high values irrespective of the demonstration type. 
One potential explanation for this discrepancy may be the inadequate training of mGPT, as these languages were insufficiently represented in its training dataset.
Furthermore, these figures demonstrates that the ratio between the translation demonstrations involving identical language pairs is minimal, suggesting that the features activated are consistent regardless of the direction of translation.
Consequently, this evidence suggests that the activation of certain features exclusively in response to the translation demonstrations is a model-independent phenomenon.\footnote{We observed consistent trends across other layers. To avoid redundancy of the main text, we have included the remaining results in the supplemental materials.}

Tab. \ref{table:mgpt_bloom_mt} presents the BLEU scores of mGPT and BLOOM.
In mGPT, the BLEU scores were very low compared to XGLM and BLOOM in overall, and the pruned models outperformed the original model.
In BLOOM, the performances of models pruned by few-shot translation demonstrations have been preserved relative to the performances of the original model before pruning. 
However, the statistical significance between the models pruned by translation demonstrations and those pruned by monolingual demonstrations was not consistently observed, as in the case of XGLM.

\section{Multilinguality of Pruned MLLMs}
\label{appendix:multilinguality}
Table \ref{table:zero-shot-all} presents the accuracy scores for each model. The table reveals that the models pruned through translation demonstrations surpass those pruned by monolingual demonstrations and the original models in performance. While LRP2  also improved the accuracy scores in XGLM and BLOOM, it did not enhance performance in mGPT. Although the scores for the models pruned using translation demonstrations are lower than those for LRP2 in BLOOM, the table suggests that this pruning strategy robustly enhances performance in non-English languages across different models, unlike LRP2.

Table \ref{table:zero-shot-marc2-xglm-7.5B} presents the accuracy scores of XGLM-7.5B, the largest model in the XGLM series, on the MARC task, and indicates that the pruning based on translation demonstrations also enhances the performance of XGLM-7.5B on non-English languages. This finding suggests the potential effectiveness of the translation demonstration-based pruning in larger-scale models.

\section{Evaluation of Cross-lingual Consistency}
\label{appendix:rankc}
We measured consistency between the predictions across languages using Ranking based Consistency (RankC) \cite{qi-etal-2023-cross}.
In this section, we explain calculation of RankC in detail.

Let us consider test examples denoted by $x_i \in X_l$ for language $l$ and $x^{\prime}_i \in X_l^{\prime}$ for another language $l^{\prime}$, along with their corresponding sets of candidate labels $\{y^1_i, ...,y^{\left| \mathcal{Y} \right|}_i \}$ and $\{y^{\prime 1}_i, ...,y^{\prime \left| \mathcal{Y} \right|}_i \}$, respectively.
Here, $y^1_i$ has the highest prediction probability and $y^{\left| \mathcal{Y} \right|}_i$ has the lowest.
RankC is defined as:

\begin{eqnarray*}
 \mathrm{RankC} (l, l^{\prime}) &=& \frac{\sum^{\left| X_l \right|}_{i=1} \mathrm{consist} (x_i, x^{\prime}_i)}{\left| X_l \right|} \\
 \mathrm{consist} (x_i, x^{\prime}_i) &=& \sum^{\left| \mathcal{Y} \right|}_{j=1} w_j \cdot P@j \\ 
 P@j &=& \frac{1}{j} \left| \{y^1_i, ...,y^j_i \} \cap \{y^{\prime 1}_i, ...,y^{\prime j}_i \} \right| \\
 w_j &=& \frac{e^{\left| \mathcal{Y} \right| - j}}{\sum^{\left| \mathcal{Y} \right|}_{k=1} e^{\left| \mathcal{Y} \right| - k}}
\label{equation:rankc}
\end{eqnarray*}

\noindent The RankC metric assign high scores when the predicted labels consistent across across languages. 

Tab. \ref{table:rankc} presents the RankC scores of each model.
This table demonstrates that the models pruned by few-shot translation demonstrations achieve the highest scores.
Therefore, the pruning using few-shot translation demonstrations improve the cross-lingual consistency of their prediction compared to the original model.

\clearpage

\begin{figure*}[t]
 \centering
 \includegraphics[scale=0.5]{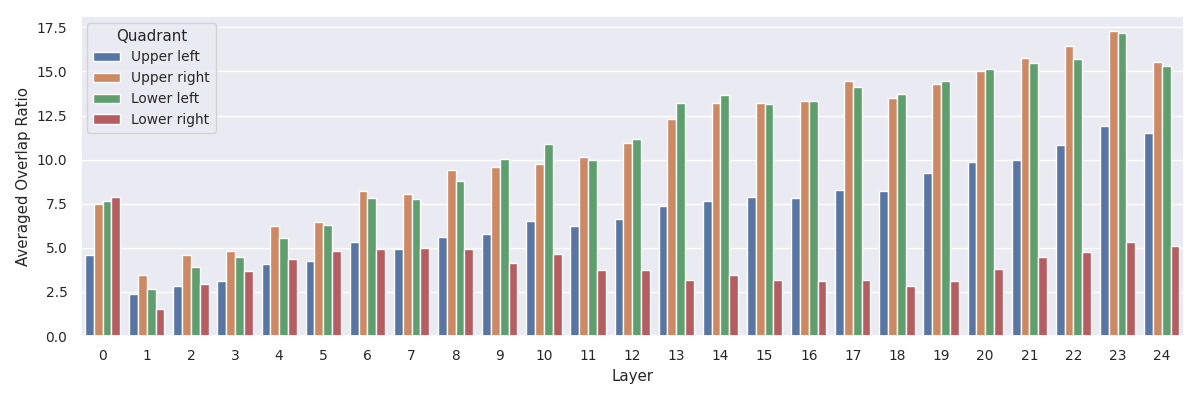}
 \caption{Averaged overlap ratios for each quadrant in mGPT.
  }\label{fig:averaged_overlap_mgpt}
\end{figure*}

\begin{figure*}[t]
 \centering
 \includegraphics[scale=0.5]{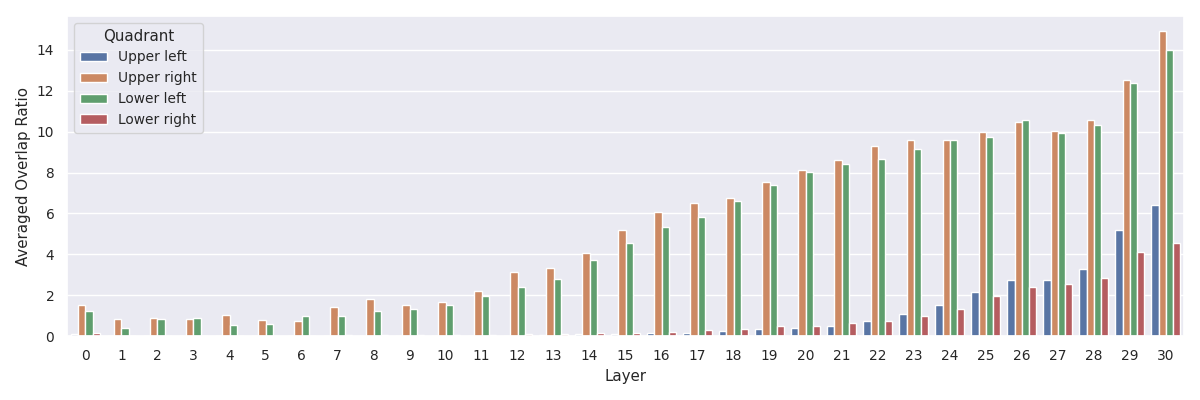}
 \caption{Averaged overlap ratios for each quadrant in BLOOM.
  }\label{fig:averaged_overlap_bloom}
\end{figure*}

\begin{table*}[tp]
\centering
\tabcolsep 4pt
\scalebox{0.65}{
\begin{tabular}{lllllllllllllllll}
\toprule
& & \multicolumn{14}{c}{Source Language} & \\
\cmidrule(lr){3-16} 
Model & Weight & Ar & Bg & De & El & Hi & Ru & Sw & Th & Tr & Ur & Vi & Fr & Es & Zh & Avg. \\
\midrule
\multirow{12}{*}{mGPT} & $\boldsymbol{\theta}$ & 3.53 & 6.14 & 8.75 & 8.76 & 1.37 & 5.16 & 3.13 & 1.55 & 2.17 & 2.62 & 5.14 & 9.16 & 9.46 & 2.02 & 4.92 \\
\cmidrule(lr){2-17} 
& $\boldsymbol{\theta}_\mathrm{En}$ & 4.42 & 6.54 & 6.10 & 9.86 & 3.00 & 4.75 & 5.28 & 4.50 & 3.63 & 2.97 & 6.17 & 6.93 & 8.36 & 5.71 & 5.58  \\
& $\boldsymbol{\theta}_\mathrm{Fr}$ & 4.35 & 7.02 & 6.98 & 10.35 & 3.24 & 4.61 & 5.35 & 4.56 & 3.19 & 2.94 & 6.15 & 6.42 & 8.39 & 5.37 & 5.63 \\
& $\boldsymbol{\theta}_\mathrm{Es}$ & 3.71 & 6.84 & 6.40 & 9.74 & 3.19 & 4.55 & 5.42 & 4.46 & 3.73 & 3.02 & 5.75 & 6.19 & 9.09 & 5.37 & 5.53 \\
& $\boldsymbol{\theta}_\mathrm{Zh}$ & 4.07 & 7.11 & 6.77 & 9.92 & 3.21 & 4.06 & 5.16 & 4.23 & 3.68 & 2.75 & 5.65 & 6.61 & 8.50 & 5.74 & 5.53 \\
& $\boldsymbol{\theta}_\mathrm{Hi}$ & 4.07 & 7.66 & 6.98 & 9.60 & 3.53 & 4.26 & 5.07 & 4.38 & 3.69 & 2.94 & 5.02 & 6.18 & 8.28 & 5.53 & 5.51 \\
& $\boldsymbol{\theta}_\mathrm{Sw}$ & 4.00 & 6.94 & 6.69 & 9.75 & 3.22 & 4.49 & 4.91 & 4.48 & 3.40 & 3.01 & 6.37 & 6.55 & 8.49 & 5.51 & 5.55 \\
\cmidrule(lr){2-17} 
& $\boldsymbol{\theta}_\mathrm{Fr-En}$ & 4.31 & 7.31$^{\dag\ddag}$ & 6.67$^{\dag}$ & 10.11$^{\dag}$ & 3.34 & 4.72 & 5.19 & 4.52 & 3.66$^{\ddag}$ & 3.19 & 5.99 & 6.30 & 8.33 & 5.77$^{\ddag}$ & 5.67 \\
& $\boldsymbol{\theta}_\mathrm{Es-En}$ & 4.23$^{\ddag}$ & 7.14$^{\dag\ddag}$ & 6.45$^{\ddag}$ & 9.72 & 3.21 & 4.94$^{\dag\ddag}$ & 5.05 & 4.36 & 3.55 & 2.96 & 6.17$^{\ddag}$ & 6.45$^{\ddag}$ & 8.19 & 5.86$^{\ddag}$ & 5.59 \\
& $\boldsymbol{\theta}_\mathrm{Zh-En}$ & 4.33$^{\ddag}$ & 7.22$^{\dag\ddag}$ & 6.38$^{\dag}$ & 9.79 & 3.32$^{\dag}$ & 4.86$^{\ddag}$ & 5.34$^{\ddag}$ & 4.39 & 3.30 & 2.94 & 6.04$^{\ddag}$ & 6.27 & 8.62$^{\dag}$ & 5.60 & 5.60 \\
& $\boldsymbol{\theta}_\mathrm{Hi-En}$ & 4.13 & 7.17$^{\dag}$ & 6.34$^{\dag}$ & 9.93 & 3.55$^{\dag}$ & 4.15 & 5.02 & 4.25 & 3.58 & 2.89 & 5.55$^{\ddag}$ & 6.39$^{\ddag}$ & 8.50$^{\ddag}$ & 5.55 & 5.50 \\
& $\boldsymbol{\theta}_\mathrm{Sw-En}$ & 3.96 & 6.99$^{\dag}$ & 6.57$^{\dag}$ & 9.46 & 3.45$^{\dag\ddag}$ & 4.49 & 5.02 & 4.68 & 3.56 & 3.01 & 5.63 & 6.33 & 8.79$^{\dag\ddag}$ & 5.57 & 5.54 \\
\midrule
\multirow{12}{*}{BLOOM} & $\boldsymbol{\theta}$ & 20.73 & 4.80 & 18.67 & 4.36 & 13.93 & 9.21 & 13.9 & 1.65 & 0.95 & 11.06 & 24.54 & 32.81 & 34.16 & 17.54 & 14.87 \\
\cmidrule(lr){2-17} 
& $\boldsymbol{\theta}^\text{Prog}_\mathrm{En}$ & 16.67 & 2.74 & 15.35 & 2.86 & 10.97 & 6.66 & 10.21 & 1.33 & 1.00 & 8.06 & 20.65 & 29.39 & 29.9 & 14.36 & 12.15\\
& $\boldsymbol{\theta}^\text{Prog}_\mathrm{Fr}$ & 20.93 & 4.14 & 18.56 & 4.32 & 13.11 & 8.9 & 14.24 & 1.48 & 1.06 & 10.76 & 24.63 & 33.01 & 33.74 & 17.04 & 14.70 \\
& $\boldsymbol{\theta}^\text{Prog}_\mathrm{Es}$ & 20.45 & 4.42 & 18.25 & 4.18 & 13.35 & 9.35 & 14.15 & 1.32 & 1.09 & 10.64 & 24.67 & 33.06 & 33.82 & 17.1 & 14.70 \\
& $\boldsymbol{\theta}^\text{Prog}_\mathrm{Zh}$ & 20.89 & 4.53 & 18.41 & 4.23 & 13.52 & 9.07 & 14.03 & 1.48 & 0.96 & 10.65 & 24.64 & 32.99 & 33.82 & 13.52 & 14.48 \\

& $\boldsymbol{\theta}^\text{Prog}_\mathrm{Hi}$ & 20.64 & 4.29 & 18.52 & 4.67 & 14.55 & 9.72 & 14.00 & 1.43 & 1.25 & 10.96 & 24.82 & 33.14 & 34.4 & 17.16 & 14.96 \\
& $\boldsymbol{\theta}^\text{Prog}_\mathrm{Sw}$ & 15.54 & 2.52 & 13.77 & 2.77 & 10.65 & 6.10 & 8.37 & 0.99 & 0.73 & 7.01 & 18.56 & 28.34 & 29.22 & 12.86 & 11.24\\
\cmidrule(lr){2-17} 
& $\boldsymbol{\theta}^\text{Prog}_\mathrm{Fr-En}$ & 21.01$^\dag$ & 4.45$^{\dag\ddag}$ & 18.53$^\dag$ & 4.00 & 13.22$^{\dag\ddag}$ & 9.12$^{\dag\ddag}$ & 13.99$^{\dag}$ & 1.59 & 1.11 & 10.69$^{\dag}$ & 24.66$^{\dag}$ & 33.98$^{\dag\ddag}$ & 33.76$^{\dag}$ & 17.10 & 14.80\\
& $\boldsymbol{\theta}^\text{Prog}_\mathrm{Es-En}$ & 20.78$^{\dag\ddag}$ & 4.43$^{\dag}$ & 18.23$^{\dag}$ & 4.12$^{\dag}$ & 13.33$^{\dag}$ & 9.54$^{\dag\ddag}$ & 14.27$^{\dag\ddag}$ & 1.54 & 1.09 & 10.64$^{\dag}$ & 24.70$^{\dag\ddag}$ & 33.24$^{\dag\ddag}$ & 33.85$^{\dag}$ & 16.88$^{\dag}$ & 14.76 \\
& $\boldsymbol{\theta}^\text{Prog}_\mathrm{Zh-En}$ & 20.77$^{\dag}$ & 4.64$^{\dag}$ & 18.74$^{\dag\ddag}$ & 4.02$^{\dag}$ & 13.30$^{\dag}$ & 9.09$^{\dag}$ & 14.31$^{\dag\ddag}$ & 1.56 & 0.97 & 10.61$^{\dag}$ & 24.85$^{\dag\ddag}$ & 32.67$^{\dag}$ & 33.85$^{\dag}$ & 17.09$^{\dag\ddag}$ & 14.74 \\
& $\boldsymbol{\theta}^\text{Prog}_\mathrm{Hi-En}$ & 20.97$^{\dag\ddag}$ & 4.33$^{\dag}$ & 18.57$^{\dag}$ & 4.57$^{\dag}$ & 14.40$^{\dag}$ & 9.65$^{\dag}$ & 14.95$^{\dag\ddag}$ & 1.38 & 1.16 & 10.95$^{\dag}$ & 24.76$^{\dag}$ & 32.87$^{\dag\ddag}$ & 34.59$^{\dag\ddag}$ & 17.83$^{\dag\ddag}$ & 15.07 \\
& $\boldsymbol{\theta}^\text{Prog}_\mathrm{Sw-En}$ & 15.92$^{\ddag}$ & 2.69 & 14.50$^{\ddag}$ & 3.00 & 11.06$^{\ddag}$ & 6.49$^{\ddag}$ & 8.63$^{\ddag}$ & 1.24 & 0.91 & 7.21 & 19.58$^{\ddag}$ & 29.20$^{\ddag}$ & 30.24$^{\dag\ddag}$ & 13.50$^{\ddag}$ & 11.72 \\
\bottomrule
\end{tabular}
}
\caption{BLEU scores on original weights and each pruned weights.
}
\label{table:mgpt_bloom_mt}
\end{table*}

\begin{table*}[tp]
\centering
\small
\tabcolsep 4pt
\scalebox{0.9}{
\begin{tabular}{llccccccccccccccc}
\toprule
Model & Weight & Ar & Bg & De & El & Hi & Ru & Sw & Th & Tr & Ur & Vi & Fr & Es & Zh & Avg. \\
\midrule
\multirow{15}{*}{XGLM} & $\boldsymbol{\theta}$ & 44.0 & 41.8& 41.6& 44.0 & 44.7& 39.5& 42.5& 44.2& 41.0 & 42.7& 45.2& 45.0 & 35.2& 43.9& 42.5 \\
\cmidrule(lr){2-17} 
& LRP2 & - & - & - & - & - & - & - & - & - & - & - &  46.4 & 36.0 & 45.1 & - \\
\cmidrule(lr){2-17} 
& $\boldsymbol{\theta}_\mathrm{En}$ & 44.6& 43.3& 42.8& 44.3& 45.0 & 41.4& 42.2& 44.1& 42.5& 42.7& 45.9& 47.1& 39.4& 46.5& 43.7 \\
& $\boldsymbol{\theta}_\mathrm{Fr}$ & 44.5 & 43.5& 41.9& 45.4& 44.6& 42.2& 42.6& 44.5& 41.7& \textbf{43.0}  & 46.9& 47.3& 38.0 & 46.6& 43.7\\
& $\boldsymbol{\theta}_\mathrm{Es}$ & 44.1& 44.0 & 42.4 & 44.3& 44.3& 42.5& 42.2& 44.6& 41.5& 42.7& 45.4& 46.8& 35.4& 46.1& 43.3\\
& $\boldsymbol{\theta}_\mathrm{Zh}$ & 44.7& 44.4& 41.5& \textbf{46.2} & \textbf{45.0} & 42.1 & 42.9 & 44.9& 42.9 & 42.2 & 46.5& 46.8& 38.5 & 46.6& 43.9 \\
& $\boldsymbol{\theta}_\mathrm{Hi}$ & 43.6 & 41.1 & 39.9 & 35.9 & 43.7 & 42.1 & 41.9 & 45.3 & 40.8 & 42.3 & 45.8 & 44.4 & 38.9 & 44.8 & 42.1 \\
& $\boldsymbol{\theta}_\mathrm{Sw}$ & 43.6 & 41.8 & 39.7 & 44.6 & 44.8 & 42.0 & 41.4 & 45.0 & 41.1 & 42.1 & 45.1 & 45.9 & 37.4 & 45.7 & 42.8 \\
\cmidrule(lr){2-17} 
& $\boldsymbol{\theta}_\mathrm{Fr-En}$ & \textbf{44.9} & 45.8 & 42.9 & \textbf{46.2} & 44.9& 43.0 & \textbf{43.5} & 45.4& 42.5& 42.8& \textbf{47.7} & 47.2& 39.7& \textbf{47.2} 
 & 44.5 \\
& $\boldsymbol{\theta}_\mathrm{Es-En}$ & 44.3 & 45.9 & 42.5& 45.6 & 44.7& 43.2& 43.3& \textbf{45.9} & 42.8& 42.5& 47.5& 47.0 & 36.3& 46.8 & 44.1 \\
& $\boldsymbol{\theta}_\mathrm{Zh-En}$ & 44.1& 45.9 & 43.0 & 45.9 & \textbf{45.0} & \textbf{43.6} & 42.5 & 45.5 & 42.2 & \textbf{43.0} & 47.5 & 47.7 & 39.8  & 46.7 & 44.4 \\
& $\boldsymbol{\theta}_\mathrm{Hi-En}$ & 43.7 & 42.8  & 40.3 & 45.2 & 44.7 & 42.5 & 42.7 & 45.5 & 41.0 & 42.2 & 45.5 & 46.2 & 37.2 & 46.0 & 43.3 \\
& $\boldsymbol{\theta}_\mathrm{Sw-En}$ & 43.8 & 44.5 & 40.3 & 46.1  & 43.7 & 41.7 & 42.0 & 45.6 & 41.7 & 42.0 & 45.4 & 46.4 & 38.9 & 46.1 & 43.4 \\

\cmidrule(lr){2-17} 
& $\boldsymbol{\theta}_\mathrm{En-Fr}$ & \textbf{44.9} & \textbf{46.8} & \textbf{43.2} & 45.9 & 44.6 & 42.7 & 43.4 & 45.7 & 42.5 & 42.9 & 47.3 & 47.6 & \textbf{40.0} & 46.8 & \textbf{44.6} \\
& $\boldsymbol{\theta}_\mathrm{En-Es}$ & 44.8 & 45.7 & 43.1 & 46.0 & 44.6 & 43.0 & 43.4 & 45.6 & \textbf{43.1} & 42.6 & 47.6 & 47.6 & 36.9 & 47.1 & 44.4 \\
& $\boldsymbol{\theta}_\mathrm{En-Zh}$ & 44.2 & 45.5 & 42.5 & 46.1 & 44.7 & 43.4 & 42.4 & 45.2 & 41.8 & 42.9 & 46.5 & \textbf{47.8} & 38.8 & 45.7 & 44.1 \\
 \midrule

\multirow{15}{*}{mGPT} & $\boldsymbol{\theta}$ & 39.2 & 39.7 & 36.0 & 41.0 & 38.9 & 39.2 & 34.0 & 43.0 & 39.9 & 39.9 & 42.2 & 42.3 & 39.4 & \textbf{41.8} & 39.7 \\
\cmidrule(lr){2-17} 
& LRP2 & - & - & - & - & - & - & - & - & - & - & -& 34.2 & 33.1 & 34.1 & - \\
\cmidrule(lr){2-17} 
& $\boldsymbol{\theta}_\mathrm{En}$ & 38.5 & 38.0 & 35.7 & 40.4 & 39.6 & 37.7 & \textbf{35.8} & 41.4 & 39.9 & 39.3 & 40.3 & 40.7 & 37.7 & 39.4 & 38.8\\
& $\boldsymbol{\theta}_\mathrm{Fr}$ & 38.4 & 39.3 & 36.0 & 42.0 & 39.5 & 38.7 & 34.3 & 42.5 & 39.8 & \textbf{40.2} & 42.5 & 42.0 & 39.0 & 40.3 & 39.6 \\
& $\boldsymbol{\theta}_\mathrm{Es}$ & 38.5 & 39.0 & 35.8 & 41.1 & 39.3 & 39.3 & 35.4 & 42.4 & 39.8 & 39.7 & 42.6 & 42.1 & 39.5 & 40.6 & 39.6 \\
& $\boldsymbol{\theta}_\mathrm{Zh}$ & 38.7 & 38.7 & 36.4 & 40.6 & 39.8 & 37.9 & 34.2 & 42.5 & 39.5 & 40.0 & 41.5 & 41.8 & 38.9 & 39.9 & 39.3\\
& $\boldsymbol{\theta}_\mathrm{Hi}$ & 39.0 & 39.0 & 34.7 & 40.8 & 40.2 & 38.3 & 34.3 & 41.6 & 38.7 & 40.6 & 42.8 & 42.1 & 38.8 & 40.3 & 39.3 \\
& $\boldsymbol{\theta}_\mathrm{Sw}$ & 38.6 & 38.1 & 34.4 & 40.4 & 39.5 & 38.2 & 34.0 & 42.1 & 39.4 & 40.2 & 43.2 & 41.7 & 39.2 & 40.7 & 39.2 \\
\cmidrule(lr){2-17} 
& $\boldsymbol{\theta}_\mathrm{Fr-En}$ & 39.3 & 39.5 & 36.3 & 41.9 & 40.2 & 39.5 & 34.7 & 42.6 & \textbf{40.2} & 40.0 & 42.8 & 42.1 & 39.7 & 41.7 & 40.0 \\
& $\boldsymbol{\theta}_\mathrm{Es-En}$ & \textbf{40.6} & \textbf{40.3} & \textbf{36.6} & \textbf{42.6} & 39.5 & \textbf{39.8} & 35.2 & 42.9 & 40.1 & 39.9 & \textbf{43.5} & \textbf{42.5} & \textbf{40.4} & 41.2 & \textbf{40.3} \\
& $\boldsymbol{\theta}_\mathrm{Zh-En}$ & 39.1 & 39.9 & 36.1 & 41.5 & 39.8 & 39.0 & 34.4 & \textbf{43.4} & 40.1 & 40.2 & 43.2 & 42.2 & 39.9 & 41.4 & 40.0 \\
& $\boldsymbol{\theta}_\mathrm{Hi-En}$  & 39.1 & 39.5 & 34.9 & 41.1 & \textbf{40.3} & 38.9 & 34.5 & 42.1 & 40.0 & 40.5 & 42.6 & 42.1 & 38.9 & 41.6 & 39.7  \\
& $\boldsymbol{\theta}_\mathrm{Sw-En}$ & 38.7 & 39.4 & 34.9 & 40.1 & 40.1 & 38.8 & 34.5 & 42.3 & 39.8 & \textbf{40.7} & 43.6 & 42.5 & 39.1 & \textbf{41.9} & 39.7 \\
\cmidrule(lr){2-17} 
& $\boldsymbol{\theta}_\mathrm{En-Fr}$ & 39.6 & 39.3 & 36.1 & 42.0 & 39.1 & 39.4 & 34.5 & 42.8 & 39.5 & 39.6 & 42.6 & 42.4 & 40.2 & 40.2 & 39.8 \\
& $\boldsymbol{\theta}_\mathrm{En-Es}$ & 39.5 & 39.7 & 36.4 & 42.0 & 39.2 & 39.1 & 35.4 & 42.4 & \textbf{40.4} & 40.2 & 42.6 & 42.2 & 39.7 & 40.6 & 40.0 \\
& $\boldsymbol{\theta}_\mathrm{En-Zh}$ & 39.4 & 39.4 & 36.2 & 42.1 & 39.7 & 39.4 & 34.5 & 42.6 & 39.3 & 40.2 & 42.3 & 42.4 & 38.8 & 40.5 & 39.8 \\
 \midrule

\multirow{26}{*}{BLOOM} & $\boldsymbol{\theta}$ & 46.7 & 40.4 & 41.9 & 38.6 & 44.9 & 40.9 & 36.8 & 36.2 & 35.9 & 41.4 & 42.9 & 45.0 & 41.1 & 45.4 & 41.2\\
\cmidrule(lr){2-17} 
& LRP2 & - & - & - & - & - & - & - & - & - & - & -& \textbf{46.0} & \textbf{44.3} & \textbf{46.8} & - \\
\cmidrule(lr){2-17} 
& $\boldsymbol{\theta}_\mathrm{En}$ & 46.4 & 40.3 & 42.5 & 39.1 & 45.2 & 41.2 & 36.8 & 35.9 & 35.6 & 41.4 & 40.9 & 44.5 & 41.2 & 43.8 & 41.0 \\
& $\boldsymbol{\theta}_\mathrm{Fr}$ & 46.4 & 40.5 & 42.2 & 39.0 & 44.9 & 40.9 & 37.1 & 35.5 & 35.1 & 41.2 & 40.9 & 44.5 & 41.5 & 44.1 & 40.9 \\
& $\boldsymbol{\theta}_\mathrm{Es}$ & 46.5 & 40.7 & 42.0 & 39.5 & 45.3 & 41.1 & 36.9 & 36 & 35.3 & 40.8 & 40.8 & 44.3 & 41.2 & 43.6 & 41.0 \\
& $\boldsymbol{\theta}_\mathrm{Zh}$ & 46.5 & 40.5 & 42.3 & 39.2 & 45.2 & 40.8 & 36.9 & 35.8 & 35.6 & 41.1 & 40.8 & 44.4 & 40.8 & 44.7 & 41.0\\
& $\boldsymbol{\theta}_\mathrm{Hi}$ & 46.3 & 40.3 & 40.2 & 40.4 & 44.5 & 41.5 & 35.5 & 35.2 & 35.5 & 41.3 & 41.4 & 43.6 & 40.6 & 43.1 & 40.6 \\
& $\boldsymbol{\theta}_\mathrm{Sw}$ & 46.8 & 40.2 & 40.6 & 40.5 & 43.7 & 41.8 & 35.5 & 35.4 & \textbf{36.1} & 41.0 & 42.3 & 44.0 & 40.8 & 44.0 & 40.9\\
\cmidrule(lr){2-17} 
& $\boldsymbol{\theta}_\mathrm{Fr-En}$ & 47.0 & 40.3 & 42.2 & 39.5 & 45.9 & 41.3 & 36.9 & 36.5 & 35.6 & 41.1 & 41.9 & 44.9 & 41.2 & 44.6 & 41.3 \\
& $\boldsymbol{\theta}_\mathrm{Es-En}$ & 47.0 & 40.6 & 42.2 & 38.9 & 45.5 & 41.1 & 37.1 & 36.6 & 35.7 & 40.9 & 41.9 & 44.5 & 41.3 & 44.4 & 41.2 \\
& $\boldsymbol{\theta}_\mathrm{Zh-En}$ & 46.7 & 40.3 & 41.9 & 39.2 & 45.3 & 41.1 & 37.0 & 36.7 & 35.7 & 40.4 & 41.6 & 44.2 & 40.5 & 45.2 & 41.1 \\
& $\boldsymbol{\theta}_\mathrm{Hi-En}$ & 47.2 & 40.7 & 40.6 & \textbf{41.1} & 45.2 & 41.6 & 36.0 & 37.0 & 36.2 & 41.4 & 42.1 & 45.4 & 42.0 & 43.7 & 41.4 \\
& $\boldsymbol{\theta}_\mathrm{Sw-En}$ & 46.9 & 40.3 & 40.6 & 40.3 & 43.8 & 41.3 & 35.7 & 36.2 & \textbf{36.1} & 41.5 & 43.0 & 44.4 & 40.8 & 44.1 & 41.0 \\
\cmidrule(lr){2-17} 
& $\boldsymbol{\theta}^\mathrm{Prog}_\mathrm{En}$ & 45.9 & 40.6 & 41.8 & 40.1 & 45.1 & 41.1 & 37.0 & 36.2 & 35.9 & 41.3 & 41.6 & 45.0 & 41.1 & 44.7 & 41.2 \\
& $\boldsymbol{\theta}^\mathrm{Prog}_\mathrm{Fr}$ & 46.9 & 40.1 & 41.7 & 39.5 & 45.8 & 40.8 & 36.8 & 35.8 & 35.2 & 40.9 & 40.8 & 44.7 & 41.2 & 45.2 & 41.1 \\
& $\boldsymbol{\theta}^\mathrm{Prog}_\mathrm{Es}$ & 47.0 & 40.1 & 41.5 & 39.7 & 46.4 & 41.0 & 36.5 & 35.8 & 35.6 & 41.1 & 41.0 & 44.4 & 42.1 & 45.1 & 41.2\\
& $\boldsymbol{\theta}^\mathrm{Prog}_\mathrm{Zh}$ & 46.8 & 40.4 & 41.5 & 39.5 & 45.2 & 40.9 & 36.6 & 35.5 & 35.3 & 41.2 & 42.3 & 45.3 & 41.8 & 45.9 & 41.3 \\
& $\boldsymbol{\theta}^\mathrm{Prog}_\mathrm{Hi}$ & 47.0 & 40.0 & 40.5 & 40.5 & 44.3 & 41.0 & 35.5 & 35.6 & 35.5 & 35.8 & 40.9 & 42.1 & 41.9 & 43.1 & 40.2  \\
& $\boldsymbol{\theta}^\mathrm{Prog}_\mathrm{Sw}$ & 46.5 & 39.5 & 38.6 & 40.1 & 45.1 & 41.2 & 35.7 & 35.6 & 35.9 & \textbf{42.5} & 42.0 & 42.9 & 39.4 & 41.6 & 40.4 \\
\cmidrule(lr){2-17} 
& $\boldsymbol{\theta}^\mathrm{Prog}_\mathrm{Fr-En}$ & 46.9 & 40.6 & 42.0 & 40.0 & 45.8 & \textbf{42.3} & \textbf{37.4} & 36.6 & 35.6 & 41.9 & 41.3 & 45.1 & 41.6 & 45.2 & 41.6 \\
& $\boldsymbol{\theta}^\mathrm{Prog}_\mathrm{Es-En}$ & 47.1 & \textbf{40.8} & 42.2 & 39.9 & \textbf{46.5} & 41.3 & 37.2 & 36.3 & 35.5 & 41.2 & 41.1 & 45.1 & 42.2 & 45.6 & 41.6 \\
& $\boldsymbol{\theta}^\mathrm{Prog}_\mathrm{Zh-En}$ & \textbf{47.2} & 40.5 & 42.2 & 40.0 & 45.8 & 41.2 & 37.1 & 36.0 & 35.8 & 41.2 & 42.1 & 45.8 & 42.4 & 46.3 & \textbf{41.7} \\
& $\boldsymbol{\theta}^\mathrm{Prog}_\mathrm{Hi-En}$ & 47.3 & 40.3 & 41.2 & 40.3 & 45.4 & 41.6 & 36.2 & 36.8 & \textbf{36.1} & 41.3 & 42.2 & 45.2 & 42.1 & 45.3 & 41.5  \\
& $\boldsymbol{\theta}^\mathrm{Prog}_\mathrm{Sw-En}$ & 46.8 & 41.1 & 42.5 & 39.5 & 45.5 & 41.2 & 36.9 & \textbf{36.6} & 35.5 & \textbf{42.5} & \textbf{43.3} & 45.2 & 41.3 & 45.0 & 41.6 \\
\cmidrule(lr){2-17} 
& $\boldsymbol{\theta}^\mathrm{Prog}_\mathrm{En-Es}$ & 46.9 & 40.3 & 42.3 & 39.8 & 46.3 & 41.0 & 37.1 & 36.2 & 35.4 & 41.3 & 41.2 & 44.9 & 41.9 & 45.2 & 41.4\\
& $\boldsymbol{\theta}^\mathrm{Prog}_\mathrm{En-Zh}$ & 46.2 & \textbf{40.8} & \textbf{42.6} & 39.9 & 45.9 & 41.3 & 36.9 & 36.1 & 35.8 & 41.0 & 42.3 & 45.7 & 42.0 & 45.5 & 41.6\\

\bottomrule
\end{tabular}
}
\caption{Accuracy scores on the XNLI task. The highest scores in each model are indicated in bold.
}
\label{table:zero-shot-all}
\end{table*}

\begin{table*}[tp]
\centering
\small
\tabcolsep 4pt
\scalebox{1.0}{
\begin{tabular}{llllllll}
\toprule
Model & Weight & De & Ja & Fr & Es & Zh & AVg. \\
\midrule
\multirow{8}{*}{XGLM-7.5B} & $\boldsymbol{\theta}$ & 76.3 & 73.0 & 73.6 & 74.7 & 74.0 & 74.3 \\
\cmidrule(lr){2-8} 
& $\boldsymbol{\theta}_\text{Fr-En}$ & 77.7$^{\dag}$ & \textbf{74.2}$^{\dag}$ & 74.8$^{\dag}$ & 75.7$^{\dag}$ & 76.6$^{\dag}$ & \textbf{75.8} \\
& $\boldsymbol{\theta}_\text{Es-En}$ & \textbf{77.8}$^{\dag}$ & 73.7$^{\dag}$ & 75.1$^{\dag}$ & 75.6$^{\dag}$ & 76.1$^{\dag}$ & 75.6 \\
& $\boldsymbol{\theta}_\text{Zh-En}$ & 77.0$^{\dag}$ & 73.9$^{\dag}$ & 75.2$^{\dag}$ & 75.8$^{\dag}$ & 76.4$^{\dag}$ & 75.6\\
& $\boldsymbol{\theta}_\text{Hi-En}$ & 76.2 & 73.1 & \textbf{75.5}$^{\dag}$ & 74.3 & \textbf{76.9}$^{\dag}$ & 75.2 \\
& $\boldsymbol{\theta}_\text{Sw-En}$ & 76.3 & 73.4$^{\dag}$ & 74.9$^{\dag}$ & \textbf{75.9}$^{\dag}$ & 75.6$^{\dag}$ & 75.2 \\
\bottomrule
\end{tabular}
}
\caption{Accuracy scores on the MARC task.}
\label{table:zero-shot-marc2-xglm-7.5B}
\end{table*}

\begin{table*}[tp]
\centering
\scalebox{0.63}{
\begin{tabular}{llccccccccccccccc}
\toprule
Model & Weight & Ar & Bg & De & El & Hi & Ru & Sw & Th & Tr & Ur & Vi & Fr & Es & Zh & Avg. \\
\midrule
\multirow{7}{*}{XGLM} & $\boldsymbol{\theta}$ & 0.790 & 0.786 & 0.441 & 0.775 & 0.520 & 0.811 & 0.824 & 0.784 & 0.770 & 0.751 & 0.791 & 0.820 & 0.792 & 0.723 & 0.741\\
\cmidrule(lr){2-17} 
& $\boldsymbol{\theta}_\mathrm{Fr}$ & 0.799 & 0.800 & 0.449 & 0.779 & 0.580 & 0.81 & 0.816 & 0.79 & 0.784 & 0.744 & 0.788 & 0.806 & 0.794 & 0.688 & 0.744 \\
& $\boldsymbol{\theta}_\mathrm{Es}$ & 0.804 & 0.795 & 0.464 & 0.79 & 0.626 & 0.811 & 0.809 & 0.793 & 0.776 & 0.764 & 0.79 & 0.811 & 0.769 & 0.683 & 0.748 \\
& $\boldsymbol{\theta}_\mathrm{Zh}$ & 0.801 & 0.785 & 0.418 & 0.785 & 0.601 & 0.814 & 0.826 & 0.802 & 0.786 & 0.770 & 0.793 & 0.819 & 0.796 & 0.694 & 0.749\\
\cmidrule(lr){2-17} 
& $\boldsymbol{\theta}_\mathrm{Fr-En}$ & 0.795 & 0.799 & 0.492 & 0.777 & 0.583 & 0.814 & 0.814 & 0.788 & 0.761 & 0.740 & 0.783 & 0.801 & 0.795 & 0.682 & 0.744 \\
& $\boldsymbol{\theta}_\mathrm{Es-En}$ & 0.801 & 0.804 & 0.494 & 0.789 & 0.604 & 0.818 & 0.816 & 0.795 & 0.773 & 0.757 & 0.790 & 0.805 & 0.774 & 0.674 & 0.749 \\
& $\boldsymbol{\theta}_\mathrm{Zh-En}$ & 0.797 & 0.803 & 0.481 & 0.785 & 0.585 & 0.823 & 0.825 & 0.802 & 0.781 & 0.755 & 0.795 & 0.804 & 0.792 & 0.681 & 0.751 \\
 \midrule

\multirow{7}{*}{mGPT} & $\boldsymbol{\theta}$ & 0.535 & 0.556 & 0.603 & 0.554 & 0.465 & 0.520 & 0.525 & 0.512 & 0.569 & 0.462 & 0.546 & 0.569 & 0.586 & 0.520 & 0.537 \\
\cmidrule(lr){2-17} 
& $\boldsymbol{\theta}_\mathrm{Fr}$ & 0.571 & 0.601 & 0.477 & 0.605 & 0.496 & 0.614 & 0.591 & 0.564 & 0.609 & 0.550 & 0.662 & 0.678 & 0.673 & 0.591 & 0.591 \\
& $\boldsymbol{\theta}_\mathrm{Es}$ & 0.589 & 0.600 & 0.485 & 0.602 & 0.493 & 0.605 & 0.573 & 0.565 & 0.599 & 0.557 & 0.653 & 0.659 & 0.658 & 0.581 & 0.587 \\
& $\boldsymbol{\theta}_\mathrm{Zh}$ & 0.595 & 0.621 & 0.461 & 0.626 & 0.481 & 0.627 & 0.581 & 0.561 & 0.626 & 0.556 & 0.670 & 0.688 & 0.698 & 0.590 & 0.598 \\
\cmidrule(lr){2-17} 
& $\boldsymbol{\theta}_\mathrm{Fr-En}$ & 0.573 & 0.593 & 0.464 & 0.610 & 0.502 & 0.624 & 0.579 & 0.581 & 0.606 & 0.549 & 0.662 & 0.673 & 0.689 & 0.603 & 0.593 \\
& $\boldsymbol{\theta}_\mathrm{Es-En}$ & 0.589 & 0.601 & 0.480 & 0.608 & 0.492 & 0.621 & 0.583 & 0.567 & 0.620 & 0.550 & 0.666 & 0.676 & 0.682 & 0.580 & 0.593 \\
& $\boldsymbol{\theta}_\mathrm{Zh-En}$ & 0.600 & 0.625 & 0.468 & 0.620 & 0.488 & 0.628 & 0.585 & 0.581 & 0.627 & 0.566 & 0.671 & 0.689 & 0.700 & 0.604 & 0.603 \\
 \midrule

\multirow{13}{*}{BLOOM} & $\boldsymbol{\theta}$ & 0.618 & 0.514 & 0.546 & 0.511 & 0.576 & 0.520 & 0.548 & 0.487 & 0.531 & 0.548 & 0.597 & 0.636 & 0.631 & 0.595 & 0.561 \\
\cmidrule(lr){2-17}
& $\boldsymbol{\theta}_\mathrm{Fr}$ & 0.613 & 0.525 & 0.560 & 0.527 & 0.572 & 0.534 & 0.546 & 0.494 & 0.533 & 0.544 & 0.595 & 0.644 & 0.630 & 0.596 & 0.565 \\
& $\boldsymbol{\theta}_\mathrm{Es}$ & 0.612 & 0.526 & 0.560 & 0.533 & 0.571 & 0.532 & 0.544 & 0.494 & 0.534 & 0.539 & 0.594 & 0.643 & 0.635 & 0.601 & 0.566 \\
& $\boldsymbol{\theta}_\mathrm{Zh}$ & 0.616 & 0.522 & 0.559 & 0.527 & 0.576 & 0.530 & 0.548 & 0.494 & 0.532 & 0.541 & 0.596 & 0.645 & 0.632 & 0.597 & 0.565 \\

\cmidrule(lr){2-17} 
& $\boldsymbol{\theta}_\mathrm{Fr-En}$ & 0.615 & 0.535 & 0.564 & 0.535 & 0.573 & 0.535 & 0.540 & 0.505 & 0.530 & 0.546 & 0.598 & 0.640 & 0.634 & 0.590 & 0.567 \\
& $\boldsymbol{\theta}_\mathrm{Es-En}$ & 0.617 & 0.534 & 0.567 & 0.535 & 0.575 & 0.538 & 0.548 & 0.500 & 0.533 & 0.542 & 0.599 & 0.642 & 0.638 & 0.596 & 0.569 \\
& $\boldsymbol{\theta}_\mathrm{Zh-En}$ & 0.617 & 0.533 & 0.564 & 0.535 & 0.574 & 0.536 & 0.544 & 0.500 & 0.534 & 0.543 & 0.598 & 0.640 & 0.634 & 0.597 & 0.568 \\

\cmidrule(lr){2-17}
& $\boldsymbol{\theta}^\mathrm{Prog}_\mathrm{Fr}$ & 0.614 & 0.539 & 0.562 & 0.532 & 0.570 & 0.534 & 0.547 & 0.507 & 0.539 & 0.547 & 0.607 & 0.645 & 0.639 & 0.592 & 0.569 \\
& $\boldsymbol{\theta}^\mathrm{Prog}_\mathrm{Es}$ & 0.618 & 0.537 & 0.564 & 0.543 & 0.568 & 0.534 & 0.555 & 0.513 & 0.549 & 0.546 & 0.616 & 0.650 & 0.651 & 0.594 & 0.574 \\
& $\boldsymbol{\theta}^\mathrm{Prog}_\mathrm{Zh}$ & 0.617 & 0.536 & 0.558 & 0.535 & 0.569 & 0.538 & 0.540 & 0.504 & 0.535 & 0.550 & 0.605 & 0.640 & 0.634 & 0.592 & 0.568 \\
\cmidrule(lr){2-17} 
& $\boldsymbol{\theta}^\mathrm{Prog}_\mathrm{Fr-En}$ & 0.624 & 0.527 & 0.560 & 0.530 & 0.570 & 0.530 & 0.554 & 0.501 & 0.540 & 0.540 & 0.607 & 0.642 & 0.642 & 0.604 & 0.569 \\
& $\boldsymbol{\theta}^\mathrm{Prog}_\mathrm{Es-En}$ & 0.626 & 0.548 & 0.564 & 0.544 & 0.568 & 0.534 & 0.557 & 0.516 & 0.542 & 0.546 & 0.610 & 0.651 & 0.655 & 0.605 & 0.576 \\
& $\boldsymbol{\theta}^\mathrm{Prog}_\mathrm{Zh-En}$ & 0.621 & 0.533 & 0.563 & 0.535 & 0.568 & 0.538 & 0.545 & 0.503 & 0.534 & 0.548 & 0.604 & 0.639 & 0.638 & 0.598 & 0.569 \\
\bottomrule
\end{tabular}
}
\caption{RankC scores between English and each language.
}
\label{table:rankc}
\end{table*}

\end{document}